September 8, 2023

# Copula Representations and Error Surface Projections for the Exclusive Or Problem

**Roy S. Freedman**[1]

**Abstract**

The exclusive or (xor) function is one of the simplest examples that illustrate why nonlinear feedforward networks are superior to linear regression for machine learning applications. We review the xor representation and approximation problems and discuss their solutions in terms of probabilistic logic and associative copula functions. After briefly reviewing the specification of feedforward networks, we compare the dynamics of learned error surfaces with different activation functions such as RELU and tanh through a set of colorful three-dimensional charts. The copula representations extend xor from Boolean to real values, thereby providing a convenient way to demonstrate the concept of cross-validation on in-sample and out-sample data sets. Our approach is pedagogical and is meant to be a machine learning prolegomenon.

**Keywords**: machine learning; neural networks; probabilistic logic; copulas; error surfaces; xor.

## 1 Introduction to the Exclusive Or Problem

The "exclusive or" (xor) function is derived from Boolean logic [1]. Figure 1.1 provides the 4-sample Boolean specification for xor and other common Boolean functions:

| | | | | | | | target |
|---|---|---|---|---|---|---|---|
| **Samples:** | *x1* | *x2* | *not(x1)* | *not(x2)* | *x1 and x2* | *x1 or x2* | *x1 xor x2* |
| *0* | 0 | 0 | 1 | 1 | 0 | 0 | 0 |
| *1* | 0 | 1 | 1 | 0 | 0 | 1 | 1 |
| *2* | 1 | 0 | 0 | 1 | 0 | 1 | 1 |
| *3* | 1 | 1 | 0 | 0 | 1 | 1 | 0 |

**Figure 1.1. The xor function together with the specification of other Boolean functions.**

Boolean inputs and outputs are restricted to the values 0 or 1. In general, for inputs $X_1$ and $X_2$, xor returns 1 when either $X_1$ or $X_2$ is 1 (but not both); otherwise xor returns 0.

### 1.1 The Exclusive Or Representation Problem

The xor representation problem is: Can we derive a formula such that, given any two Boolean inputs, the formula exactly computes the xor output? Here is one formula that represents xor in terms of the logic functions shown in Figure 1.1:

$$X_1 \text{ xor } \mathrm{X}_2 = \left(X_1 \text{ or } X_2\right) \text{and} \left(\text{not}\left(\left(X_1 \text{ and } X_2\right)\right)\right).$$

[1] Roy S. Freedman is with Inductive Solutions, Inc., New York, NY 10280 and with the Department of Finance and Risk Engineering, New York University Tandon School of Engineering, Brooklyn NY 11201. Email: roy@inductive.com xor roy.freedman@nyu.edu.

It is easy to validate this expression with the table in Figure 1.1. Let's now see if we can find a simple formula based on arithmetic operations. Let's start with a linear formula. In general, an arbitrary linear formula for two inputs is $\text{out}(x_1, x_2) = w_1 \cdot x_1 + w_2 \cdot x_2 + w_3$, with unknowns $w_1, w_2, w_3$ unrestricted real numbers (called "weights"). Inputs $x_1, x_2$ are (so far) restricted to the sample input values of 0 or 1. If a linear formula exists, we need to show specific values for these three unknown weights.

Let's evaluate $\text{out}(x_1, x_2)$ on each Boolean sample. From Figure 1.1, we have $\text{out}(0,0) = \text{out}(1,1) = 0$, and $\text{out}(0,1) = 1$ $\text{out}(1,0) = 1$. After substitution we have:

$$\begin{cases} \text{out}(0,0) = 0 = w_1 \cdot 0 + w_2 \cdot 0 + w_3 = w_3 \\ \text{out}(0,1) = 1 = w_1 \cdot 0 + w_2 \cdot 1 + w_3 = w_2 \\ \text{out}(1,0) = 1 = w_1 \cdot 1 + w_2 \cdot 0 + w_3 = w_1 \\ \text{out}(1,1) = 0 = w_1 \cdot 1 + w_2 \cdot 1 + w_3 = w_1 + w_2 \end{cases}$$

There are four linear equations but three unknowns: this system is overdetermined. Note that the first equation implies $w_3 = 0$; the second implies $w_2 = 1$ and the third implies $w_1 = 1$. Finally, the last equation implies $w_1 + w_2 = 0$. But since we already determined that $w_1 = 1$ and $w_2 = 1$, this means that $2 = 0$ which is impossible: our original assumption that a linear formula $\text{out}(x_1, x_2) = w_1 \cdot x_1 + w_2 \cdot x_2 + w_3$ exists for xor is false! .

## 1.2 The Exclusive Or Approximation Problem

The xor approximation problem is: Can we derive a formula that, given any two Boolean inputs, the formula "reasonably approximates" the xor output? Note that "reasonable" is subjective.

The conventional machine learning (and statistical) solution is to postulate a family of functions and pick a member of the family with a best "goodness-of-fit" measure. Without loss of generality, we choose our goodness-of-fit measure as the sum (over all samples) of the squares of the errors.

Let's see how this looks for a few examples. Consider five candidate functions for the xor approximation:

$$\text{F}_a(x_1, x_2) := 1 \qquad \text{F}_b(x_1, x_2) := 0 \qquad \text{F}_c(x_1, x_2) := 0.5$$
$$\text{F}_d(x_1, x_2) := 2 \cdot x_1 + 2 \cdot x_2 - 1 \qquad \text{F}_e(x_1, x_2) := x_1 + x_2 - 2 \cdot (x_1 \cdot x_2)$$

The first four formulas $\text{F}_a - \text{F}_d$ are linear formulas; the last formula is nonlinear because of the $(x_1 \cdot x_2)$ term. Figure 1.2 shows the values of the outputs of these candidates over the four samples.

| | | | *targ* | | | | | |
|---|---|---|---|---|---|---|---|---|
| **Samples:** | *x1* | *x2* | *xor* | Fa | Fb | *Fc* | *Fd* | *Fe* |
| *0* | 0 | 0 | 0 | 1 | 0 | 0.5 | -1 | 0 |
| *1* | 0 | 1 | 1 | 1 | 0 | 0.5 | 1 | 1 |
| *2* | 1 | 0 | 1 | 1 | 0 | 0.5 | 1 | 1 |
| *3* | 1 | 1 | 0 | 1 | 0 | 0.5 | 3 | 0 |

**Figure 1.2. Candidate solutions $\text{F}_a - \text{F}_d$ to the xor approximation problem: results over all samples.**

Formula $F_a$ gives the exact answers for Sample 1 and Sample 2 but gives the wrong answers for Sample 0 and Sample 3. In the language of binary classification (as used in diagnostic tests), the outputs for Sample 0 and Sample 3 are false positives; the outputs for Sample 1 and Sample 2 are true positives.

Formula $F_b$ gives the exact answers for Sample 0 and Sample 3 but gives the wrong answers for Sample 1 and Sample 2. In the language of binary classification, the outputs for Sample 0 and Sample 3 are true negatives; the outputs for Sample 1 and Sample 2 are false negatives.

Formula $F_c$ gives the wrong answers for all samples. Moreover, we now need to interpret the concept of a false positive or false negative for non-Boolean outputs that are neither 0 nor 1.

Formula $F_d$ gives the exact answers for Sample 1 and Sample 2 (true positives) but gives the wrong answers for Sample 0 and Sample 3. Moreover, the results seem "out-of-bounds" in some sense. Suppose we modify the outputs by introducing a procedure for rounding: if $F_d(x_1, x_2) < 1$ then set $\text{out}(x_1, x_2) = 0$ otherwise we set $\text{out}(x_1, x_2) = 1$. In this case we have the correct answers for the first three samples and a false positive for the last sample.

Nonlinear Formula $F_e$ gives the exact answers for all samples.

Let's see how these candidates perform with using the sum (over all samples) of the squares of the errors as the goodness-of-fit measure. For any (exact) solution to the xor representation problem like $F_e$, note that the sum of the squares of the errors over all samples is zero.

This is not the case for our candidate solutions $F_a - F_d$. For $F_a$, the sum of the squares of the errors is $(0-1)^2 + (1-1)^2 + (1-1)^2 + (0-1)^2 = 2$; for $F_b$, the sum of the squared errors also evaluates to $(0-0)^2 + (1-0)^2 + (1-0)^2 + (0-0)^2 = 2$. For formula $F_c$ the sum of the squares of the errors is $(0-0.5)^2 + (1-0.5)^2 + (1-0.5)^2 + (0-0.5)^2 = 1$. So in this sense, $F_c$ provides a better "goodness-of-fit" than either $F_a$ and $F_b -$ even though $F_c$ gives the wrong answers for all samples. (Is this "reasonable?") Figure 1.3 shows each sample error (in columns *SqErr*) and the goodness-of-fit for $F_a - F_e$ .

| | | | *targ* | | | | | | | | | | |
|---|---|---|---|---|---|---|---|---|---|---|---|---|---|
| **Samples:** | *x1* | *x2* | *xor* | Fa | *SqErr* | Fb | *SqErr* | *Fc* | *SqErr* | *Fd* | *SqErr* | *Fe* | *SqErr* |
| *0* | 0 | 0 | 0 | 1 | *1* | 0 | *0* | 0.5 | *0.25* | -1 | *1* | 0 | *0* |
| *1* | 0 | 1 | 1 | 1 | *0* | 0 | *1* | 0.5 | *0.25* | 1 | *0* | 1 | *0* |
| *2* | 1 | 0 | 1 | 1 | *0* | 0 | *1* | 0.5 | *0.25* | 1 | *0* | 1 | *0* |
| *3* | 1 | 1 | 0 | 1 | *1* | 0 | *0* | 0.5 | *0.25* | 3 | *9* | 0 | *0* |
| | | **SUMSqErrors:** | | | **2** | | **2** | | **1** | | **10** | | ***0*** |

**Figure 1.3. Squares of the errors SqErr over all samples of $F_a - F_e$ and the resultant goodness-of-fit (sum of the squares of the errors SUMSqErrors).**

Let's return to the simple linear formula $\text{out}(x_1, x_2) = w_1 \cdot x_1 + w_2 \cdot x_2 + w_3$ and ask: can we find a linear function that has the best "goodness-of-fit?" In the family of linear functions, we need to find weights $w_1, w_2, w_3$ that minimizes

$$
\begin{aligned}
\text{err}(w_1, w_2, w_3) &= \left(\text{out}(0,0) - 0\right)^2 + \left(\text{out}(0,1) - 1\right)^2 + \left(\text{out}(1,0) - 1\right)^2 + \left(\text{out}(1,1) - 0\right)^2 \\
&= \left(w_3\right)^2 + \left(w_1 + w_2 + w_3\right)^2 + \left(w_1 - 1 + w_3\right)^2 + \left(-1 + w_2 + w_3\right)^2
\end{aligned}
$$

The following procedure solves this minimization problem. Calculus tells us we need to find the critical points of $\text{err}(w_1, w_2, w_3)$. To do this, compute the gradient – the slopes (first partial derivatives) at an arbitrary point $w_1, w_2, w_3$; the critical points are where to the slopes are all zero (in our case this corresponds to finding the unique bottom of a quadratic 3-dimensional bowl-like surface in 4-dimensional space). The result is three equations in the three unknowns $w_1, w_2, w_3$:

$$\begin{cases} 4 \cdot w_1 + 2 \cdot w_2 + 4 \cdot w_3 - 2 = 0 \\ 2 \cdot w_1 + 4 \cdot w_2 + 4 \cdot w_3 - 2 = 0 \\ 4 \cdot w_1 + 4 \cdot w_2 + 8 \cdot w_3 - 4 = 0 \end{cases}$$

This system has solution $w_1 = 0;\ w_2 = 0;\ w_3 = 0.5$. For the family of linear functions, $\mathrm{F}_c$ is the best we can do. No other combination of weights yields a goodness-of-fit smaller than 1. The solution to the xor approximation problem for the family of linear functions is $\mathrm{F}_c$. This procedure is essentially the method used in linear regression (such as found in Excel's LINEST spreadsheet function).

Note that we can do better in the universe of nonlinear functions like $\mathrm{F}_e$. We can also bring some nonlinear functions to the linear world. For the xor problem, suppose we add a third input column $x_3$, where for all samples we define $x_3 = x_1 \cdot x_2$; the new $x_3$ column values are (0,0,0,1). The new xor problem is to represent xor by $\text{out}(x_1, x_2, x_3) = w_1 \cdot x_1 + w_2 \cdot x_2 + w_3 \cdot x_3 + w_4$, now with goodness-of-fit a function of four weights: $\text{err}(w_1, w_2, w_3, w_4)$. The linear regression procedure yields $w_1 = 1;\ w_2 = 1;\ w_3 = -2;\ w_4 = 0$ so that $\mathrm{F}_g(x_1, x_2, x_3) = x_1 + x_2 - 2 \cdot x_3$. This formula is an exact representation of xor that corresponds to $\mathrm{F}_e$. Incorporating more inputs in this manner frequently improves results. But in general, we do not know a priori, which new input to add.

### 1.3 Exclusive Or, Rounding, and Linear Discriminants

What about other nonlinear functions? Note that nonlinear functions include expressions involving products, exponentials, hyperbolic and trigonometric functions, and so on; they also involve piecewise continuous functions (using operations like rounding to the nearest whole number).

Let's provide an example of a rounding procedure. First apply linear regression to the "and" function and "or" function described in Figure 1. The results are:

$$\begin{cases} \text{R_and}(x_1, x_2) := 0.5 \cdot x_1 + 0.5 \cdot x_2 - 0.25 \\ \text{R_or}(x_1, x_2) := 0.5 \cdot x_1 + 0.5 \cdot x_2 + 0.25 \end{cases}$$

Note that the outputs of the regression are extended from Boolean 1 and 0 to real "analog" numbers between 0 and 1. The rounding procedure is: if the result is greater than 0.5 then round up and return 1; otherwise round down and return 0. Rounding enables us to exactly represent the "and" function and "or" function with Boolean outputs by

$$\begin{cases} \text{outAnd}(x_1, x_2) = \text{if } \ (0.5 \cdot x_1 + 0.5 \cdot x_2 - 0.25) > 0.5 \text{ then return 1, else return 0.} \\ \text{outOr}(x_1, x_2) = \text{if } \ (0.5 \cdot x_1 + 0.5 \cdot x_2 + 0.25) > 0.5 \text{ then return 1, else return 0.} \end{cases}$$

This representation is called a linear discriminant. We use linear regression to find the optimal weights and introduce a rounding procedure at the last step. Figure 1.4 shows the results that solve the representation problem for “or” and “and”.

Note that discriminants implicitly extends the Boolean inputs (as well as outputs) to real analog numbers between 0 and 1. For example, let’s evaluate the discriminants on a few analog inputs: $\text{outOr}(0.2,0)=0$; $\text{outOr}(0.2,0.4)=1$; $\text{outAnd}(0.2,0.4)=0$; $\text{outAnd}(0.8,0.9)=1$.

| Samples: | x1 | x2 | and | or | R_and | outAnd | R_or | outOr |
|---|---|---|---|---|---|---|---|---|
| 0 | 0 | 0 | 0 | 0 | -0.25 | 0 | 0.25 | 0 |
| 1 | 0 | 1 | 0 | 1 | 0.25 | 0 | 0.75 | 1 |
| 2 | 1 | 0 | 0 | 1 | 0.25 | 0 | 0.75 | 1 |
| 3 | 1 | 1 | 1 | 1 | 0.75 | 1 | 1.25 | 1 |

**Figure 1.4. Linear discriminants.**

For both analog inputs between zero and one (denoted by $0 \le x_1, x_2 \le 1$), the regression functions $\text{R_and}(x_1,x_2)$ and $\text{R_or}(x_1,x_2)$ separate the real $(x_1,x_2)$ analog space into different regions, depending whether the value is greater or less than the threshold 0.5. Figure 1.5 shows this graphically for $\text{R_or}(x_1,x_2)$ and $\text{outOr}(x_1,x_2)$.

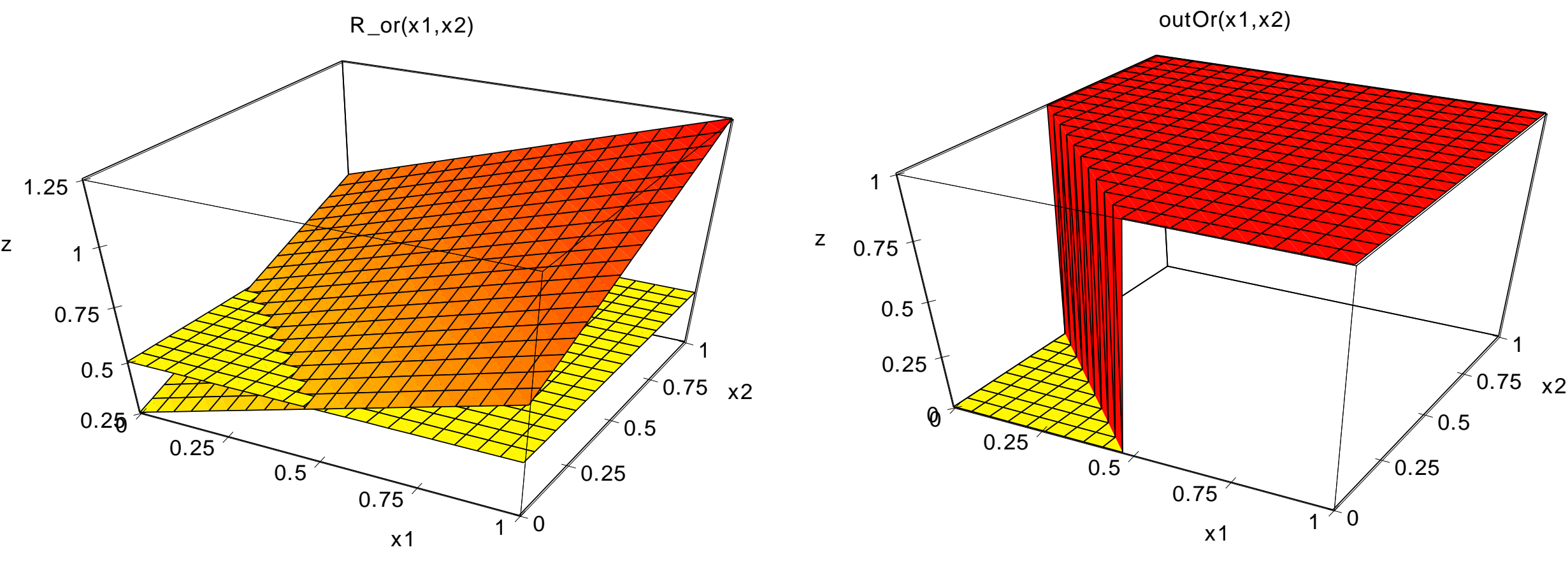

**Figure 1.5. Linear separability. Left: Regression function for “or” showing 0.5 threshold as flat plane. All (x1,x2) points whose regression values are above 0.5 are classed as “1”; otherwise as “0”. Right: the linear discriminant for “or”. The corners agree with the “or” specification (Figure 1.1).**

Note that this discriminant procedure does not work for the xor problem: the linear regression result $\text{R_xor}(x_1,x_2)=\text{F}_c(x_1,x_2)=0.5$ is a constant! There is no discriminating formula that can be used for any rounding. This is why (in machine learning jargon), the xor problem is not linearly separable [2].

## 1.4 Outline of Approach

It turns out that for other families of nonlinear functions, we can derive results that solve both xor representation and approximation problems. Section 2 introduces probabilistic logic and the nonlinear copula functions of probability theory. Probabilistic logic is used in artificial intelligence to model uncertainty in a way that is consistent with both probability and logic [3-4]: it provides a consistent (probabilistic) interpretation of inputs and outputs as probabilities having values between 0 and 1. Copulas are the natural continuous extensions of the "and" Boolean function to continuous analog functions having inputs and outputs between 0 and 1 [5-6]. We show how certain copulas solve the xor representation problem.

Section 3 reviews the representation of linear function families using array and matrix notation. We show how cascading feedforward formulas – where the outputs of one formula are the inputs to another formula – are simply represented by "neural" networks of array operations. We show why adding additional layers of network complexity to linear families does not add any value in representational capability. In Section 4 we extend our representation to nonlinear families. Section 5 shows how these representations work with a standard machine learning techniques (backpropagation): we create simple nonlinear networks that solve the xor approximation problem. The networks learn copula and non-copula representations. Section 6 shows examples of the xor error surfaces for the network under a variety of function families. The shapes of these error surfaces provide a nice insight as to why backpropagation succeeds (or fails) in finding the "best" weights based on an initial starting point on the surface. Finally, Section 7 discusses the machine learning principles of cross-validation in determining whether it is "reasonable" to conclude if the xor approximation converges to an xor copula representation.

## 2 Probabilistic Logic and Copula Function Representations of Xor

Boolean expressions correspond to propositions: statements that are either TRUE (represented as 1) or FALSE (represented as 0). This correspondence of TRUE and FALSE to Boolean values is extensively used in the design of digital switching circuits [1]. For many machine learning applications, we frequently extend Boolean values to real (analog) values between zero and one. We saw how this can be useful with inputs to linear discriminants: we use the analog output of a linear regression with a rounding procedure that brings us back to the Boolean world. In this section our approach is different: we allow both inputs and outputs to be analog: real numbers between zero and one. We usually interpret these values as a degree of plausibility or probability. If we do this in a way that is consistent both with Boolean logic and probability theory, then the result is probabilistic logic [3].

### 2.1 Probabilistic Logic

We focus on sets of statements (Boolean expressions). Given a set of statements, the set of statements across all consistent truth value assignments specifies a discrete sample space. Probabilistic logic assigns a probability to this sample space (called the Nilsson space in [4]). The probability of a particular statement is the frequency of values where the statement is TRUE (viz, having value 1) under all consistent assignments following the rules of Boolean logic.

This is easy to see with an example. Let's give a probabilistic interpretation to the table in Figure 1.1. Denote statements having Boolean values by upper class letters and probabilities by lower case letters. The table shows the set of all consistent assignments of TRUE (having value 1) and FALSE (having value 0) to statements: $X_1$, $X_2$, $\text{not}(X_1)$, $\text{not}(X_2)$, $X_1 \text{ and } X_2$, $X_1 \text{ or } X_2$, $X_1 \text{ xor } X_2$. The sample space in Figure 1.1 is the set of 4 truth value assignments to $(X_1, X_2)$ : all possible Boolean inputs which determine the values of the other output Boolean functions. The frequency of values where a statement is TRUE shows, for example, $\Pr[X_1] = \frac{2}{4}$; $\Pr[X_2] = \frac{2}{4}$; $\Pr[X_1 \text{ and } \mathrm{X}_2] = \frac{1}{4}$; $\Pr[X_1 \text{ or } \mathrm{X}_2] = \frac{3}{4}$; $\Pr[X_1 \text{ xor } \mathrm{X}_2] = \frac{2}{4}$.

This is not the only possible assignment. Use the sample space of Figure 1.1 to randomly select a Boolean sample assignment $(x_1, x_2)$ together with the corresponding values (consistent with Boolean rules) of the output functions. A set of 10 random samples is shown below in Figure 2.1 for: $X_1$, $X_2$ and three corresponding functions: $X_1$ and $X_2$, $X_1$ or $X_2$, $X_1$ xor $X_2$. (This set of random assignments correspond to a random data selection used in machine learning.) We can use this random sample to compute the empirical frequencies – an estimate of probabilities. In Figure 2.1, these data show $\Pr[X_1] = \frac{5}{10}$; $\Pr[X_2] = \frac{5}{10}$; $\Pr[X_1 \text{ and } X_2] = \frac{3}{10}$; $\Pr[X_1 \text{ or } X_2] = \frac{7}{10}$; $\Pr[X_1 \text{ xor } X_2] = \frac{4}{10}$.

| Sample | x1 | x2 | *and* | *or* | *xor* | x1 | x2 | *and* | *or* | *xor* | Sample |
|---|---|---|---|---|---|---|---|---|---|---|---|
| **0** | 0 | 0 | 0 | 0 | 0 | 1 | 0 | 0 | 1 | 1 | **5** |
| **1** | 0 | 0 | 0 | 0 | 0 | 0 | 0 | 0 | 0 | 0 | **6** |
| **2** | 1 | 0 | 0 | 1 | 1 | 1 | 1 | 1 | 1 | 0 | **7** |
| **3** | 1 | 1 | 1 | 1 | 0 | 1 | 1 | 1 | 1 | 0 | **8** |
| **4** | 0 | 1 | 0 | 1 | 1 | 0 | 1 | 0 | 1 | 1 | **9** |

**Figure 2.1. 10 random samples of Boolean data for several Boolean functions.**

The four requirements for a probabilistic logic that assures that probabilities on Boolean statements $X$ and $Y$ are consistent with classical probability theory are:

1. $0 \le \Pr[X] \le 1$.
2. If $X$ is always TRUE then $\Pr[X] = 1$. If $X$ is always False then $\Pr[X] = 0$.
3. $\Pr[X \text{ or } Y] = \Pr[X] + \Pr[Y] - \Pr[X \text{ and } Y]$.
4. $\Pr[\text{not}(X)] = 1 - \Pr[X]$.

These requirements are used to prove [4] the following important inequalities:

$$0 \le \Pr[X \text{ and } Y] \le \min\left(\Pr[X], \Pr[Y]\right)$$
$$\max\left(\Pr[X], \Pr[Y]\right) \le \Pr[X \text{ or } Y] \le 1$$

Any probability measure consistent with probabilistic logic satisfies these upper and lower bounds for “and” and “or”. To simplify our notation: let $x = \Pr[X]$ so $0 \le x \le 1$. For $x = \Pr[X]$ and $y = \Pr[Y]$, define functions $A(x, y)$ and $R(x, y)$ by:

$$A(x, y) = \Pr[X \text{ and } Y] = A\left(\Pr[X], \Pr[Y]\right)$$
$$R(x, y) = \Pr[X \text{ or } Y] = R\left(\Pr[X], \Pr[Y]\right)$$

Note that $0 \le A(x, y) \le 1$ and $0 \le R(x, y) \le 1$.

By the four requirements for probabilistic logic, $A(x, y)$ and $R(x, y)$ satisfy the following “And/Or Boundary Conditions.” These four conditions specify the values of $A(x, y)$ and $R(x, y)$ for $x = 0$, $x = 1$; $y = 0$, $y = 1$, corresponding to the probability of the statements $X$ and FALSE, $X$ and TRUE, $X$ or FALSE, $X$ or TRUE:

$$\Pr[\mathit{FALSE}\text{ and }X] = A(0,x) = 0 = A(x,0) = \Pr[X\text{ and }\mathit{FALSE}] = \Pr[\mathit{FALSE}] = 0$$
$$\Pr[\mathit{TRUE}\text{ and }X] = A(1,x) = x = A(x,1) = \Pr[X\text{ and }\mathit{TRUE}] = \Pr[X];$$
$$\Pr[\mathit{FALSE}\text{ or }X] = R(0,x) = x = R(x,0) = \Pr[X\text{ or }\mathit{FALSE}] = \Pr[X];$$
$$\Pr[\mathit{TRUE}\text{ or }X] = R(1,x) = 1 = R(x,1) = \Pr[X\text{ or }\mathit{TRUE}] = \Pr[\mathit{TRUE}] = 1.$$

Note that the four Boolean combinations for $(x, y)$ – $(0,0)$, $(0,1)$, $(1,0)$, $(1,1)$ – specify the Boolean "and" and "or" functions in terms of probabilities. For example:

For $\Pr[X]=0$ and $\Pr[Y]=0$: $\Pr[X\text{ and }Y] = A(\Pr[X],\Pr[Y]) = A(0,0) = 0$;

For $\Pr[X]=0$ and $\Pr[Y]=1$: $\Pr[X\text{ and }Y] = A(\Pr[X],\Pr[Y]) = A(0,1) = 0$;

For $\Pr[X]=1$ and $\Pr[Y]=0$: $\Pr[X\text{ and }Y] = A(\Pr[X],\Pr[Y]) = A(1,0) = 0$;

For $\Pr[X]=1$ and $\Pr[Y]=1$: $\Pr[X\text{ and }Y] = A(\Pr[X],\Pr[Y]) = A(1,1) = 1$.

Similarly: $R(0,0)=0$; $R(0,1)=1=R(1,0)$; $R(1,1)=1$. By the third requirement for probabilistic logic, we have:

$$A(x, y) + R(x, y) = x + y$$

So, given $A(x, y)$ we can always find $R(x, y)$; given $R(x, y)$ we can always find $A(x, y)$. Consequently: we can always evaluate the probability of any Boolean expression formed with "and", "or", and "not" with $A(x, y)$ and $R(x, y)$. Let's see what this looks like for xor. From the formula given in Section 1:

$$\begin{aligned}
\Pr[X\text{ xor }Y] &= \Pr\Big[(X\text{ or }Y)\text{and}\big(\text{not}((X\text{ and }Y))\big)\Big] \\
&= \Pr[X\text{ or }Y] + \big(1-\Pr[X\text{ and }Y]\big) - \Pr\Big[(X\text{ or }Y)\text{or}\big(\text{not}((X\text{ and }Y))\big)\Big] \\
&= \Pr[X\text{ or }Y] + \big(1-\Pr[X\text{ and }Y]\big) - \Pr\Big[(X\text{ or }Y)\text{or}\big(\text{not}(X)\big)\text{or}\big(\text{not}(Y)\big)\Big] \\
&= \Pr[X\text{ or }Y] - \Pr[X\text{ and }Y]
\end{aligned}$$

Thus, the probability of $X\text{ xor }Y$ is, for $x=\Pr[X]$ and $y=\Pr[Y]$:

$$\boxed{F(x, y) = \Pr[X\text{ xor }Y] = R(x, y) - A(x, y) = x + y - 2\cdot A(x, y).}$$

### 2.2 Associative (and Commutative) Copulas

Since "and" and "or" are associative and commutative, probabilistic logic implies the probabilities formed by association and commutation are the same:

$$\Pr\left[X \text{ and} \left(Y \text{ and} Z\right)\right] = \Pr\left[\left(X \text{ and} Y\right) \text{and} Z\right];$$
$$\Pr\left[X \text{ or} \left(Y \text{ or } Z\right)\right] = \Pr\left[\left(X \text{ or } Y\right) \text{ or } Z\right];$$
$$\Pr\left[X \text{ and} Y\right] = \Pr\left[Y \text{ and} X\right];$$
$$\Pr\left[X \text{ or } Y\right] = \Pr\left[Y \text{ or} X\right].$$

Let $W = \left(Y \text{ and} Z\right)$ and $V = \left(X \text{ and} Y\right)$. Since the probabilities of association are equal,

$$\Pr\left[X \text{ and} W\right] = A\left(\Pr\left[X\right], \Pr\left[W\right]\right) = A\left(\Pr\left[X\right], \Pr\left[Y \text{ and} Z\right]\right) = A\left(\Pr\left[X\right], A\left(\Pr\left[Y\right], \Pr\left[Z\right]\right)\right)$$
$$\Pr\left[V \text{ and} Z\right] = A\left(\Pr\left[V\right], \Pr\left[Z\right]\right) = A\left(\Pr\left[X \text{ and} Y\right], \Pr\left[Z\right]\right) = A\left(A\left(\Pr\left[X\right], \Pr\left[Y\right]\right), \Pr\left[Z\right]\right)$$

So $\Pr\left[X \text{ and} W\right] = \Pr\left[V \text{ and} Z\right]$ implies $A\left(x, A\left(y,z\right)\right) = A\left(A\left(x,y\right),z\right)$: $A(x, y)$ is associative.

$A(x, y)$ is also commutative, since $A(x, y) = \Pr\left[X \text{ and} Y\right] = \Pr\left[Y \text{ and} X\right] = A(y, x)$. Similarly, $R(x, y)$ is associative and commutative.

It turns out that these properties completely specify a function family. Frank [5] proved the following result: Let $A(x, y)$ and $R(x, y)$ be real-valued functions defined on $0 \le x_1, x_2 \le 1$ so that $0 \le A(x_1, x_2), R(x_1, x_2) \le 1$. Suppose they are (a) both continuous and associative; (b) are related by $A(x, y) + R(x, y) = x + y$; and (c) satisfy the four conditions:

$$A(0, x) = 0 = A(x, 0); \; A(1, x) = x = A(x, 1);$$
$$R(0, x) = x = R(x, 0); \; R(1, x) = 1 = R(x, 1).$$

Then the only functions $A(x, y)$, $R(x, y)$ that satisfy (a-c) have the parameterized representations $A(x, y) = \mathrm{A}_s(x_1, x_2)$ where:

For $s \in (0, \infty), s \ne 1$: $\quad \mathrm{A}_s(x_1, x_2) = \log_s \left( 1 + \dfrac{\left(s^{x_1} - 1\right) \cdot \left(s^{x_2} - 1\right)}{s - 1} \right)$

when $s \to 0$: $\quad \mathrm{A}_0(x_1, x_2) = \min(x_1, x_2)$

when $s \to 1$: $\quad \mathrm{A}_1(x_1, x_2) = x_1 \cdot x_2$

when $s \to \infty$: $\quad \mathrm{A}_\infty(x_1, x_2) = \max(x_1 + x_2 - 1, 0)$

Function $\mathrm{A}_s(x, y)$ is an example of an Archimedean copula [6]. In probability theory, copula functions correspond to representations of a joint density of a set of arbitrary random variables in terms of a joint density of a set of uniformly distributed random variables. In the context of probabilistic logic, copulas provide a consistent way to extend a set of Boolean (0,1) values to a set of analog values between 0 and 1. If the Boolean samples (inputs and outputs) are consistent with probabilistic logic, then the probability of a statement represented as a Boolean function can represented with operations involving $\mathrm{A}_s(x, y)$ and $\mathrm{R}_s(x, y)$. Note the inequalities:

$$\mathrm{A}_\infty(x_1,x_2) \le \mathrm{A}_1(x_1,x_2) \le \mathrm{A}_0(x_1,x_2) \text{ so that } \mathrm{A}_\infty(x_1,x_2) \le \mathrm{A}_s(x_1,x_2) \le \mathrm{A}_o(x_1,x_2) \text{ for } 0 \le s \le \infty.$$

For the xor representation problem, we have from above that Frank's parameterization implies:

$$F(x,y) = \mathrm{F}_s(x_1,x_2) := \mathrm{R}_s(x_1,x_2) - \mathrm{A}_s(x_1,x_2) = x_1 + x_2 - 2\cdot \mathrm{A}_s(x_1,x_2).$$

The four conditions (c) imply

$$\mathrm{F}_s(x,0) = \mathrm{F}_s(0,x) = x \text{ and } \mathrm{F}_s(x,1) = \mathrm{F}_s(1,x) = 1 - x\ .$$

Thus, the only representations of xor that is consistent with probabilistic logic, is $\mathrm{F}_s(x_1,x_2)$:

$$\mathrm{F}_0(x_1,x_2) = x_1 + x_2 - 2\cdot\min(x_1,x_2) = \max(x_1,x_2) - \min(x_1,x_2);$$
$$\mathrm{F}_1(x_1,x_2) = x_1 + x_2 - 2\cdot x_1\cdot x_2;$$
$$\mathrm{F}_s(x_1,x_2) = x_1 + x_2 - 2\cdot\log_s\left(1 + \frac{(s^{x_1}-1)\cdot(s^{x_2}-1)}{s-1}\right),\ 0 < s < \infty,\ s \ne 1;$$
$$\mathrm{F}_\infty(x_1,x_2) = x_1 + x_2 - 2\cdot\max(x_1+x_2-1,0) = \min(x_1+x_2,1) - \max(x_1+x_2-1,0)\ .$$

Note that for $\mathrm{F}_s(x_1,x_2)$, the xor inequalities are:

$$\mathrm{F}_0(x_1,x_2) \le \mathrm{F}_1(x_1,x_2) \le \mathrm{F}_\infty(x_1,x_2) \text{ so that } \mathrm{F}_0(x_1,x_2) \le \mathrm{F}_s(x_1,x_2) \le \mathrm{F}_\infty(x_1,x_2) \text{ for } 0 \le s \le \infty.$$

Figure 2.2 shows three dimensional charts of $\mathrm{F}_0(x_1,x_2)$, $\mathrm{F}_1(x_1,x_2)$, and $\mathrm{F}_\infty(x_1,x_2)$: the left sides of the figure plots $(x_1,x_2,z)$ with height $z = \mathrm{F}_s(x_1,x_2)$; the right side plots the three dimensional contours (the lines where xor has a constant value). Figure 2.3 shows three dimensional charts for $s$=0.01; 0.5; 0.75; 1.25; 2, 20. These formulas are continuous functions defined for all $(x_1,x_2)$ between 0 and 1: they also agree with the xor specification at the 4 corner sample points $\mathrm{F}_s(0,0) = 0 = \mathrm{F}_s(1,1); \mathrm{F}_s(0,1) = \mathrm{F}_s(1,0) = 1$. They continuously interpolate xor. Note that $\mathrm{F}_0$ and $\mathrm{F}_\infty$ both have sharp corners and sharp folds: this implies the gradients (slopes) at those corners and folds are not continuous.

Note that these copula functions have multiple representations. To see this, recall the Heaviside unit step function $u(t)$ which can be defined by: $u(t) = 1, t > 0$ and 0 for $t \le 0$. Note that

$$x = x\cdot u(x) + x\cdot u(-x)$$
$$|x| = x\cdot u(x) + (-x)\cdot u(-x)$$
$$x + y = \big(x\cdot u(y-x) + x\cdot u(x-y)\big) + \big(y\cdot u(y-x) + y\cdot u(x-y)\big)$$
$$\min(x,y) = x\cdot u(y-x) + y\cdot u(x-y)$$

So, for example, $x + y - \min(x,y) = x\cdot u(x-y) + y\cdot u(y-x) = \max(x,y)$. This implies

$$\mathrm{F}_0(x_1,x_2) := \max(x_1,x_2) - \min(x_1,x_2) = |x_1 - x_2|\ .$$

**Figure 2.2. Copula solutions to the xor representation problem: s=0; s=1; s=∞.**

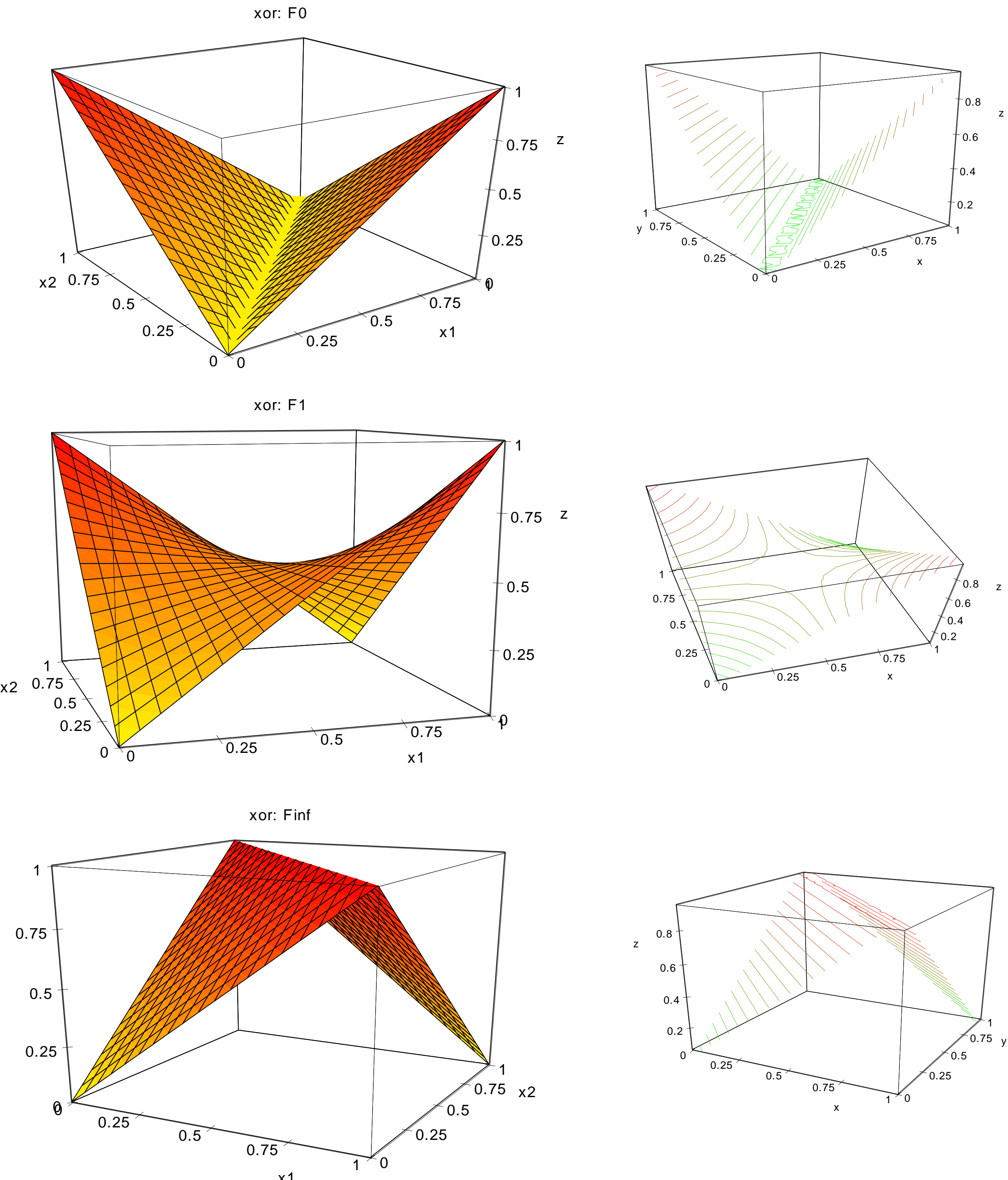

**Figure 2.3. Copula solutions to the xor representation problem for various values of s.**

xor: Fs; s=0.01

xor: Fs; s=0.5

xor: Fs; s=0.75

xor: Fs; s=1.25

xor: Fs; s=2

xor: Fs; s=20

### 2.3 Representations of xor with Frank's Copulas

Let's review the truth assignments of Figure 1.1 and 2.1 to examine consistency with the probabilistic inequalities and Frank's associative copula. The Figure 1.1 assignment is consistent with the probabilistic logic inequalities:

$$\Pr\left[X_1 \text{ and } \mathrm{X}_2\right] = 0.25 \le \min\left(\Pr\left[X_1\right], \Pr\left[X_2\right]\right) = \min\left(0.5, 0.5\right) = 0.5\,;$$
$$0.5 = \max\left(0.5, 0.5\right) = \max\left(\Pr\left[X_1\right], \Pr\left[X_2\right]\right) \le \Pr\left[X_1 \text{ or } \mathrm{X}_2\right] = 0.75\,.$$

Also note that according to Figure 1.1, $\Pr\left[X_1\right] = \Pr\left[X_2\right] = 0.5$ and $\Pr\left[X_1 \text{ and } \mathrm{X}_2\right] = 0.25$. When we solve for $s$ in $\mathrm{A}_s\left(0.5, 0.5\right) = 0.25$, we find that $s$=1 in Frank's representation. The xor probability evaluates to $\mathrm{F}_1\left(x_1, x_2\right) = (\frac{1}{2} + \frac{1}{2}) - 2 \cdot \frac{1}{4} = 0.5 = \Pr\left[X_1 \text{ xor } \mathrm{X}_2\right]$.

The Figure 2.1 assignment is also consistent with the probabilistic logic inequalities:

$$\Pr\left[X_1 \text{ and } \mathrm{X}_2\right] = 0.3 \le \min\left(\Pr\left[X_1\right], \Pr\left[X_2\right]\right) = \min\left(0.5, 0.5\right) = 0.5\,;$$
$$0.5 = \max\left(0.5, 0.5\right) = \max\left(\Pr\left[X_1\right], \Pr\left[X_2\right]\right) \le \Pr\left[X_1 \text{ or } \mathrm{X}_2\right] = 0.7\,.$$

According to Figure 2.1, $\Pr\left[X_1\right] = \Pr\left[X_2\right] = 0.5$ and $\Pr\left[X_1 \text{ and } \mathrm{X}_2\right] = 0.3$. When we solve for $s$ in $\mathrm{A}_s\left(0.5, 0.5\right) = 0.3$, we find that $s = 0.193$ in Frank's representation. The xor probability evaluates to $\mathrm{F}_s\left(0.5, 0.5\right) = (\frac{1}{2} + \frac{1}{2}) - 0.6 = 0.4 = \Pr\left[X_1 \text{ xor } \mathrm{X}_2\right]$. For another example, consider the table of Figure 2.4.

| Sample | x1 | x2 | *and* | *or* | *xor* | x1 | x2 | *and* | *or* | *xor* | Sample |
|---|---|---|---|---|---|---|---|---|---|---|---|
| **0** | 1 | 0 | 0 | 1 | 1 | 1 | 1 | 1 | 1 | 0 | **5** |
| **1** | 0 | 1 | 0 | 1 | 1 | 1 | 0 | 0 | 1 | 1 | **6** |
| **2** | 1 | 0 | 0 | 1 | 1 | 0 | 1 | 0 | 1 | 1 | **7** |
| **3** | 0 | 1 | 0 | 1 | 1 | 0 | 1 | 0 | 1 | 1 | **8** |
| **4** | 0 | 1 | 0 | 1 | 1 | 0 | 1 | 0 | 1 | 1 | **9** |

**Figure 2.4. Another set of 10 random samples from the sample space of Figure 1.1.**

The empirical frequencies are $\Pr\left[X_1\right] = \frac{4}{10}$; $\Pr\left[X_2\right] = \frac{7}{10}$; $\Pr\left[X_1 \text{ and } \mathrm{X}_2\right] = \frac{1}{10}$; $\Pr\left[X_1 \text{ or } \mathrm{X}_2\right] = 1$. The assignment is consistent with the probabilistic logic inequalities:

$$\Pr\left[X_1 \text{ and } \mathrm{X}_2\right] = 0.1 \le \min\left(\Pr\left[X_1\right], \Pr\left[X_2\right]\right) = \min\left(0.4, 0.7\right) = 0.4\,;$$
$$0.7 = \max\left(0.4, 0.7\right) = \max\left(\Pr\left[X_1\right], \Pr\left[X_2\right]\right) \le \Pr\left[X_1 \text{ or } \mathrm{X}_2\right] = 1\,.$$

When we solve for $s$ in $\mathrm{A}_s\left(0.4, 0.7\right) = 0.1$ we find that this probability assignment corresponds to $s = \infty$ in Frank's representation.: $\mathrm{A}_\infty\left(0.4, 0.7\right) = \max(0.4 + 0.7 - 1, 0) = 0.1$.

## 3 Array Representations of Linear Families

This next two sections review an array notation which helps us account for most functional families used in machine learning. As with all accounting notations it is easier to see with an example: these are provided in Figures 3.1-3.4.

First, recall the equation for a line: $y = a \cdot x + b$. Here, $a$ is the slope and $b$ is the y-intercept. For two inputs (as in the linear regression examples for xor discussed in Sections 1.2 and 1.3), we have seen how this generalizes to formulas of the form

$$y := w_{11} \cdot x_1 + w_{12} \cdot x_2 + b$$

Using matrix multiplication (denoted by $\bullet$ by mathematicians and MMULT in Excel spreadsheets), we write this as

$$\begin{aligned} \mathbf{y} &= \mathbf{A} \bullet \mathbf{x} + \mathbf{b} \\ &= \begin{pmatrix} w_{11} & w_{12} \end{pmatrix} \bullet \begin{pmatrix} x_1 \\ x_2 \end{pmatrix} + (b) \\ &= \text{MMULT(A,x)+b} \end{aligned}$$

Here, $\mathbf{x}$ is a column array and $\mathbf{A}$ is a row array. We can subsume the constant $\mathbf{b}$ further with matrix multiplication. Create a column array $\mathbf{inp}$ by adding an extra row to $\mathbf{x}$ and a weight array $\mathbf{w}$ by adding an extra column to $\mathbf{A}$:

$$\mathbf{inp} := \begin{pmatrix} x_1 \\ x_2 \\ 1 \end{pmatrix}.$$

Then the output is

$$\text{out} := \mathbf{w} \bullet \mathbf{inp} = \begin{pmatrix} w_{11} & w_{12} & w_{13} \end{pmatrix} \bullet \begin{pmatrix} x_1 \\ x_2 \\ 1 \end{pmatrix} = w_{11} \cdot x_1 + w_{12} \cdot x_2 + w_{13} \cdot 1 = w_{11} \cdot x_1 + w_{12} \cdot x_2 + w_{13} = \mathbf{y}.$$

So $\mathbf{inp}$ has an extra row assigned to 1 and $\mathbf{w}$ has an additional component where $w_{13} = b$. Let's apply this to the xor problem. Represent each sample $\mathbf{x}$ by its own column array $\mathbf{inp}$:

$$\mathbf{inp}(0) := \begin{pmatrix} 0 \\ 0 \\ 1 \end{pmatrix}; \mathbf{inp}(1) := \begin{pmatrix} 0 \\ 1 \\ 1 \end{pmatrix}; \mathbf{inp}(2) := \begin{pmatrix} 1 \\ 0 \\ 1 \end{pmatrix}; \mathbf{inp}(3) := \begin{pmatrix} 1 \\ 1 \\ 1 \end{pmatrix}.$$

Group them into single array:

$$\mathbf{Inputs} := \begin{pmatrix} 0 & 0 & 1 & 1 \\ 0 & 1 & 0 & 1 \\ 1 & 1 & 1 & 1 \end{pmatrix}.$$

Similarly, group the xor target outputs for all samples in a row array: **Targets** $:= \begin{pmatrix} 0 & 1 & 1 & 0 \end{pmatrix}$. Figure 3.1 shows the original **Samples** for the xor problem, the array **Inputs**, and the array **Targets**.

| | A | B | C | D | E | F |
|---|---|---|---|---|---|---|
| 2 | | | | | | |
| 3 | | **Samples:** | *x1* | *x2* | *xor* | |
| 4 | | *0* | 0 | 0 | 0 | |
| 5 | | *1* | 0 | 1 | 1 | |
| 6 | | *2* | 1 | 0 | 1 | |
| 7 | | *3* | 1 | 1 | 0 | |
| 8 | | | | | | |
| 9 | | **Inputs:** | *inp(0)* | *inp(1)* | *inp(2)* | *inp(3)* |
| 10 | | *x1* | 0 | 0 | 1 | 1 |
| 11 | | *x2* | 0 | 1 | 0 | 1 |
| 12 | | | 1 | 1 | 1 | 1 |
| 13 | | **Targets:** | | | | |
| 14 | | *xor* | 0 | 1 | 1 | 0 |
| 15 | | | | | | |
| 16 | | **Weights:** | *w1* | *w2* | *w3* | |
| 17 | | | 1.11E-16 | 1.11E-16 | 0.5 | |
| 18 | | | | | | |

**Figure 3.1. Spreadsheet arrays for the xor representation problem for the linear family.**

For the linear family, it can be shown that the weights with the best goodness-of-fit measure (using the sum of the squares of the errors) is given by the array formula:

```
=TRANSPOSE( MMULT( MINVERSE( MMULT(INPUTS,TRANSPOSE(INPUTS))),
        MMULT(INPUTS,TRANSPOSE(TARGETS) )  )  )
```

This formula uses the spreadsheet array functions MMULT, MINVERSE, and TRANSPOSE. In Figure 3.1, the following formula is entered in the **Weights** array C17:E17.

```
=TRANSPOSE( MMULT( MINVERSE( MMULT(C10:F12,TRANSPOSE(C10:F12))),
        MMULT(C10:F12,TRANSPOSE(C14:F14) )  )  )
```

Cell E17 shows the weight w3=0.5; the other weights are negligible. Variations of this formula are used to compute the weights in linear regression.

Let's introduce a complication by allowing outputs of one formula to be inputs to another. This feedforward cascade of inputs to outputs to inputs defines "layers" of nested computation. For example, given inputs $x_1$ and $x_2$, first compute $y_1$ and $y_2$ and then use these to compute the final output:

$$
\begin{aligned}
y_1 &:= w_{11}^1 \cdot x_1 + w_{12}^1 \cdot x_2 + w_{13}^1 \\
y_2 &:= w_{11}^1 \cdot x_1 + w_{12}^1 \cdot x_2 + w_{13}^1 \\
\text{out} &:= w_{11}^2 \cdot x_1 + w_{12}^2 \cdot x_2 + w_{13}^2
\end{aligned}
$$

The formulas are easier to see with array notation. From inputs **inp** to outputs, we introduce the intermediate arrays **y** and **out1**.

The first input is:

$$\mathbf{inp} := \begin{pmatrix} x_1 \\ x_2 \\ 1 \end{pmatrix};$$

We have as intermediate outputs:

$$\mathbf{y} := \mathbf{w}^1 \bullet \mathbf{inp} = \begin{pmatrix} w_{11}^1 & w_{12}^1 & w_{13}^1 \\ w_{21}^1 & w_{22}^1 & w_{23}^1 \end{pmatrix} \bullet \begin{pmatrix} x_1 \\ x_2 \\ 1 \end{pmatrix} = \begin{pmatrix} w_{11}^1 \cdot x_1 + w_{12}^1 \cdot x_2 + w_{13}^1 \\ w_{21}^1 \cdot x_1 + w_{22}^1 \cdot x_2 + w_{23}^1 \end{pmatrix} = \begin{pmatrix} y_1 \\ y_2 \end{pmatrix};$$

The second input (also referred to as the "input of the first layer") is built from the first intermediate output:

$$\mathbf{out1} := \begin{pmatrix} y_1 \\ y_2 \\ 1 \end{pmatrix}$$

The final (second layer) output is then:

$$\text{out} := \mathbf{w}^2 \bullet \mathbf{out1} = \begin{pmatrix} w_{11}^2 & w_{12}^2 & w_{13}^2 \end{pmatrix} \bullet \begin{pmatrix} \mathbf{y} \\ 1 \end{pmatrix} = \begin{pmatrix} w_{11}^2 & w_{12}^2 & w_{13}^2 \end{pmatrix} \bullet \begin{pmatrix} y_1 \\ y_2 \\ 1 \end{pmatrix} = w_{11}^2 \cdot y_1 + w_{12}^2 \cdot y_2 + w_{13}^2 .$$

A full evaluation yields:

$$\begin{aligned} \text{out} := \mathbf{w}^2 \bullet \mathbf{out1} &= w_{11}^2 \cdot y_1 + w_{12}^2 \cdot y_2 + w_{13}^2 \\ &= w_{11}^2 \cdot \left( w_{11}^1 \cdot x_1 + w_{12}^1 \cdot x_2 + w_{13}^1 \right) + w_{12}^2 \cdot \left( w_{21}^1 \cdot x_1 + w_{22}^1 \cdot x_2 + w_{23}^1 \right) + w_{13}^2 \end{aligned}$$

This is easier to see on a spreadsheet. Let's visualize this in Excel: define cell regions **inp** (the inputs C10:C12); **w1_** and **w2_** (two weight regions D10:F11 and H10:J10); and **out1_** and **out** (the two output regions G10:G12 and K10); and install the array formulas in **out1_** and **out** as shown:

| | B | C | D | E | F | G | H | I | J | K | L |
|---|---|---|---|---|---|---|---|---|---|---|---|
| 9 | | **inp** | **w1_** | | | **out1_** | **w2_** | | | **out** | |
| 10 | | *x_1* | *w1_11* | *w1_12* | *w1_13* | =MMULT(**w1_**, **inp**) | *w2_11* | *w2_12* | *w2_13* | =MMULT(**w2_** ,**out1** ) | |
| 11 | | *x_2* | *w1_21* | *w1_22* | *w1_23* | =MMULT(**w1_**, **inp**) | | | | | |
| 12 | | 1 | | | | 1 | | | | | |
| 13 | | | | | | | | | | | |

In mathematical notation, the superscript denotes the weight matrix number, subscript denotes the row-column coordinates so $w_{21}^1$ in mathematical notation corresponds to `w1_21` in programming (or spreadsheet) notation. To see examples of computation, populate the weight arrays with some random values: the results of the formulas using these random weights for **out1**_and **out** are:

| inp | w_1 | | | out_1 | w_2 | | | out |
|---|---|---|---|---|---|---|---|---|
| 0 | 0.1 | -0.1 | 0.2 | 0.1 | -0.4 | -0.2 | 0.3 | 0.18 |
| 1 | -0.2 | 0.3 | 0.1 | 0.4 | | | | |
| 1 | | | | 1 | | | | |

The array formulas in the cells are displayed here:

| inp | w_1 | | | out_1 | w_2 | | | out |
|---|---|---|---|---|---|---|---|---|
| 0 | 0.1 | -0.1 | 0.2 | =MMULT(w_1,inp) | -0.4 | -0.2 | 0.3 | =MMULT(w_2,out_1) |
| 1 | -0.2 | 0.3 | 0.1 | =MMULT(w_1,inp) | | | | |
| 1 | | | | 1 | | | | |

The resultant cascading feedforward network has topology "2-2-1", corresponding to two (x1,x2) inputs (2-), two intermediate computed (y1,y2) outputs (-2-), and one final computed output **out** (-1 for the last layer). The number of intermediate outputs is also referred to as "the number of hidden layer units in the neural network." Note that the initial inputs are not included in the layer count. The network structure can be viewed by using the "Trace Dependents" spreadsheet tool; the results are seen here:

| inp | w_1 | | | out_1 | w_2 | | | out |
|---|---|---|---|---|---|---|---|---|
| 0 | 0.1 | -0.1 | 0.2 | 0.1 | -0.4 | -0.2 | 0.3 | 0.18 |
| 1 | -0.2 | 0.3 | 0.1 | 0.4 | | | | |
| 1 | | | | 1 | | | | |

Does adding layers change the representational power of the linear family? Lets's see by adding layers to this 2-2-1 network.

For the first layer, specify the following arrays for the first set of weights:

$$\mathbf{w}^1 = \begin{pmatrix} w_{11}^1 & w_{12}^1 & w_{13}^1 \\ w_{21}^1 & w_{22}^1 & w_{23}^1 \end{pmatrix} = \begin{pmatrix} \mathbf{A}^1 & \mathbf{b}^1 \end{pmatrix} \text{ so } \mathbf{A}^1 = \begin{pmatrix} w_{11}^1 & w_{12}^1 \\ w_{21}^1 & w_{22}^1 \end{pmatrix}; \quad \mathbf{b}^1 = \begin{pmatrix} w_{13}^1 \\ w_{23}^1 \end{pmatrix}.$$

For the second layer, specify the following arrays for the second set of weights:

$$\mathbf{w}^2 = \begin{pmatrix} w_{11}^2 & w_{12}^2 & w_{13}^2 \end{pmatrix} = \begin{pmatrix} \mathbf{A}^2 & \mathbf{b}^2 \end{pmatrix} \text{ so } \mathbf{A}^2 = \begin{pmatrix} w_{11}^2 & w_{12}^2 \end{pmatrix}; \quad \mathbf{b}^2 = \begin{pmatrix} w_{13}^2 \end{pmatrix}$$

Then

$$\mathbf{y} = \mathbf{w}^1 \bullet \mathbf{inp} = \mathbf{A}^1 \bullet \mathbf{x} + \mathbf{b}^1 \quad \text{(first layer computation)}$$
$$\mathbf{out} = \mathbf{w}^2 \bullet \mathbf{out1} = \mathbf{A}^2 \bullet \mathbf{y} + \mathbf{b}^2 \text{ (second layer computation).}$$

Consequently, by substituting the outputs for the inputs, we obtain:

$$\mathbf{out} = \mathbf{A}^2 \bullet \mathbf{y} + \mathbf{b}^2 = \mathbf{A}^2 \bullet \left(\mathbf{A}^1 \bullet \mathbf{x} + \mathbf{b}^1\right) + \mathbf{b}^2 = \left(\mathbf{A}^2 \bullet \mathbf{A}^1\right) \bullet \mathbf{x} + \left(\mathbf{A}^2 \bullet \mathbf{b}^1 + \mathbf{b}^2\right)$$

Thus, the 2-2-1 network can be represented as a 2-1 (single layer) linear equivalent, with the weights defined by

$$\mathbf{w} = \left( \mathbf{A}^2 \bullet \mathbf{A}^1 \quad \mathbf{A}^2 \bullet \mathbf{b}^1 + \mathbf{b}^2 \right)$$

Consequently,

$$\mathbf{out} = \mathbf{w} \bullet \mathbf{inp} = \left( w_{11} \quad w_{12} \quad w_{13} \right) \bullet \mathbf{inp} = \left( \mathbf{A} \quad \mathbf{b} \right) \bullet \mathbf{inp} = \left( \mathbf{A}^2 \bullet \mathbf{A}^1 \quad \mathbf{A}^2 \bullet \mathbf{b}^1 + \mathbf{b}^2 \right) \bullet \mathbf{inp}$$

It does not matter how many layers cascade in linear network family: they are equivalent to a single layer. The 2-n-1 is same as the 2-1 topology. Applying this rule recursively: a 2-n-m-1 network is the same as a 2-n-1 topology which is the same as a 2-1 topology. (This result generalizes for any number of inputs since the number of initial inputs is not specified in the matrix multiplications.)

As before, this is easiest to see on a spreadsheet: see Figure 3.2.

| inp | w1_ | A1_ | b1_ | out1 | w2_ | A2_ | b2_ | out |
|---|---|---|---|---|---|---|---|---|
| 0 | 0.1 | -0.1 | 0.2 | 0.1 | -0.4 | -0.2 | 0.3 | 0.18 |
| 1 | -0.2 | 0.3 | 0.1 | 0.4 | | | | |
| 1 | | | | 1 | | | | |

| inp | w1_ | A1_ | b1_ | out1_ | w2_ | A2_ | b2_ | out |
|---|---|---|---|---|---|---|---|---|
| 0 | 0.1 | -0.1 | 0.2 | =MMULT(w1_,inp) | -0.4 | -0.2 | 0.3 | =MMULT(w2_,out_1) |
| 1 | -0.2 | 0.3 | 0.1 | =MMULT(w1_,inp) | | | | |
| 1 | | | | 1 | | | | |

| inp | w_ | A_ | b_ | out |
|---|---|---|---|---|
| 0 | 0 | -0.02 | 0.2 | 0.18 |
| 1 | | | | |
| 1 | | | | |

| inp | w_ | A_ | b_ | out |
|---|---|---|---|---|
| 0 | =MMULT(A2_,A1_) | =MMULT(A2_,A1_) | =MMULT(A2_,b1_)+b2_ | =MMULT(w_,inp) |
| 1 | | | | |
| 1 | | | | |

**Figure 3.2. Moving from top to bottom: (1) a 2-2-1 network showing output values and weight arrays A1_ and A2_ (in gray) and b1_ and b2_ (yellow); (2) the formula view of the 2-2-1 network; (3) the equivalent 2-1 network showing the new A_ (gray) and new b_ (yellow); (4) the formulas of the equivalent 2-1 network showing the new weight formulas and new output formulas.**

## 4 Array Representations of Nonlinear Families

Here we introduce nonlinearities into the feedforward cascading linear networks described in the previous section. Let $f(r)$ denote a real valued function, sometimes specified by $f : \mathbb{R} \to \mathbb{R}$.

What happens when we evaluate a real-valued function on an array? The simplest solution is to evaluate the function on every element in the array:

$$f(\mathbf{x}) = f\begin{bmatrix} x_1 \\ x_2 \end{bmatrix} = \begin{bmatrix} f(x_1) \\ f(x_2) \end{bmatrix}$$

In functional programming, this is called "mapping" a function: applying a function to the components of the array. This is similar to the situation of multiplying an array by a number: we just multiply every element in the array by the number:

$$r \cdot \mathbf{x} = r \cdot \begin{bmatrix} x_1 \\ x_2 \end{bmatrix} = \begin{bmatrix} r \cdot x_1 \\ r \cdot x_2 \end{bmatrix}$$

This array multiplication is sometimes called scalar multiplication. For example, for real-valued function $f(x) = \tanh(x)$,

$$f\begin{bmatrix} x_1 \\ x_2 \end{bmatrix} = \tanh\begin{bmatrix} x_1 \\ x_2 \end{bmatrix} = \begin{bmatrix} \tanh(x_1) \\ \tanh(x_2) \end{bmatrix}.$$

In machine learning jargon, these mapping functions are called "activation functions." We show some examples in Figure 4.1; [7] has a nice table of activation functions used in neural networks.

| Function: Math/Excel | Name |
|---|---|
| $f(t) = \tanh(t)$ <br> `=tanh(t)` | Hyperbolic tangent |
| $f(t) = 1/(1+\exp(-t))$ <br> `=1/(1+exp(-t))` | Sigmoid |
| $f(t) = Id(t)$ <br> `=t` | Identity function |
| $f(t) = t \cdot u(t)$ <br> `=if(t>0,t,0)` | Rectified Linear Unit (RELU) |

**Figure 4.1. Some Real-Valued Activation Functions.**

Mapping a function is a very special case of applying a general nonlinear function on an array. In general, a nonlinear function that takes pairs of real numbers and returns pairs of real numbers is denoted by $g : \mathbb{R}^2 \to \mathbb{R}^2$. These functions are sometimes called "vector-valued functions." An example of such a function is

$$g\left(\begin{bmatrix} x_1 \\ x_2 \end{bmatrix}\right) = \begin{bmatrix} \tanh(x_1 \cdot x_2) + x_1 \cdot x_2 \\ x_2 / \left(1 + \exp(-x_1)\right) \end{bmatrix}$$

It is much easier to study mapping of real-valued activation functions then the more general vector valued nonlinear functions.

Let's see how activation function maps look with the 2-1 family. All we need to do is apply the real-valued activation function to the linear outputs. Here is a nonlinear output:

$$\text{out} := f(w_{11} \cdot x_1 + w_{12} \cdot x_2 + w_{13}) \,.$$

Note that for the identity function $f(t) = Id(t) = t$, this nonlinear family reduces to the linear family since in this case $\text{out} := Id(w_{11} \cdot x_1 + w_{12} \cdot x_2 + w_{13}) = w_{11} \cdot x_1 + w_{12} \cdot x_2 + w_{13}$.

Next, let's look at the nonlinear 2-2-1 network family. Apply function $f_1$ to the intermediate linear outputs to obtain:

$$\mathbf{y} = f_1\left(\mathbf{A^1} \bullet \mathbf{x} + \mathbf{b^1}\right) = f_1\left(\mathbf{w}^1 \bullet \mathbf{inp}\right) = \begin{pmatrix} f_1\left(w^1_{11} \cdot x_1 + w^1_{12} \cdot x_2 + w^1_{13}\right) \\ f_1\left(w^1_{21} \cdot x_1 + w^1_{22} \cdot x_2 + w^1_{23}\right) \end{pmatrix} = \begin{pmatrix} y_1 \\ y_2 \end{pmatrix}$$

The second input is built from the first intermediate outputs:

$$\mathbf{out1} := \begin{pmatrix} y_1 \\ y_2 \\ 1 \end{pmatrix}$$

The final output is then derived by applying a function $f_2$ to the final linear output:

$$\begin{aligned} \text{out} &:= f_2\left(\mathbf{w}^2 \bullet \mathbf{out1}\right) = f_2\left(\begin{pmatrix} w^2_{11} & w^2_{12} & w^2_{13} \end{pmatrix} \bullet \begin{pmatrix} \mathbf{y} \\ 1 \end{pmatrix}\right) \\ &= f_2\left(\begin{pmatrix} w^2_{11} & w^2_{12} & w^2_{13} \end{pmatrix} \bullet \begin{pmatrix} y_1 \\ y_2 \\ 1 \end{pmatrix}\right) = f_2\left(w^2_{11} \cdot y_1 + w^2_{12} \cdot y_2 + w^2_{13}\right) \end{aligned}$$

A full evaluation yields:

$$\boxed{\begin{aligned} \text{out} &:= f_2\left(\mathbf{w}^2 \bullet \mathbf{out1}\right) = f_2\left(w^2_{11} \cdot y_1 + w^2_{12} \cdot y_2 + w^2_{13}\right) \\ &= f_2\left(w^2_{11} \cdot \left(f_1\left(w^1_{11} \cdot x_1 + w^1_{12} \cdot x_2 + w^1_{13}\right)\right) + w^2_{12} \cdot \left(f_1\left(w^1_{21} \cdot x_1 + w^1_{22} \cdot x_2 + w^1_{23}\right)\right) + w^2_{13}\right) \end{aligned}}$$

Note that this nonlinear network output reduces to the linear network output (see the bottom of page 16 in Section 3) when all activation functions are set to the identity function: $f_1 = f_2 = Id$. However, in general, the nonlinear 2-2-1 network cannot be simplified.

In analogy to the network topology notation (i.e., "2-2-1") we introduce a network activation notation. For example, "inp-f1-f2" means that we use f1 as the activation function of the first layer, and f2 as the activation function of the second (final output) layer.

Let's visualize this in Excel. Define cell regions **inp** (the inputs C10:C12); **w1_** and **w2_** (two weight regions D10:F11 and H10:J10); and **out1** and **out** (the two output regions G10:G12 and K10), as shown:

| | B | C | D | E | F | G | H | I | J | K |
|---|---|---|---|---|---|---|---|---|---|---|
| 9 | | **inp** | **w1_** | | | **out1** | **w2_** | | | **out** |
| 10 | | *x_1* | *w1_11* | *w1_12* | *w1_13* | =f1( MMULT(**w1_**, **inp**) ) | *w2_11* | *w2_12* | *w2_13* | =f2( MMULT(**w2_**,**out1** ) ) |
| 11 | | *x_2* | *w1_21* | *w1_22* | *w1_23* | =f1( MMULT(**w1_**, **inp**) ) | | | | |
| 12 | | 1 | | | | 1 | | | | |
| 13 | | | | | | | | | | |

This 2-2-1/inp-f1-f2 specification looks very similar to the 2-2-1 network of linear functions we discussed in Section 3. In our notation, the linear network specified by 2-2-1 is identical to by 2-2-1/inp-f1-f2 for f1=f2=Id. The network structure can be viewed by using the Trace Dependents spreadsheet tool; the results are seen in Figure 4.1 showing 2-2-1/inp-tanh-tanh.

| **inp** | **w1_** | | | **out1** | **w2_** | | | **out** |
|---|---|---|---|---|---|---|---|---|
| 0 | *0.1* | *-0.1* | *0.2* | 0.099667995 | *-0.4* | *-0.2* | *0.3* | 0.182089511 |
| 1 | *-0.2* | *0.3* | *0.1* | 0.379948962 | | | | |
| 1 | | | | 1 | | | | |

| **inp** | **w1_** | | | **out1** | **w2_** | | | **out** |
|---|---|---|---|---|---|---|---|---|
| 0 | *0.1* | *-0.1* | *0.2* | 0.099667995 | *-0.4* | *-0.2* | *0.3* | 0.182089511 |
| 1 | *-0.2* | *0.3* | *0.1* | 0.379948962 | | | | |
| 1 | | | | 1 | | | | |

**Figure 4.1. 2-2-1 topology showing output cell dependency network, random values for weights, for network 2-2-1/inp-tanh-tanh.**

Nonlinear networks of the type where the nonlinearity is created by mapping real-valued activation functions have "universal approximation capabilities" as described by [8-10]. Essentially this means that just as we can approximate any real number (like $\pi$ or $\sqrt{2}$ ) by a fraction or decimal to any degree of accuracy, so we can approximate any bounded continuous function by a nonlinear feedforward network with a sufficiently large number of intermediate outputs. This implies that any bounded continuous function with two inputs can be approximated by a 2-n-1/inp-f1-f2 network with n sufficiently large. Just as the rational numbers are dense on the real line, so are these nonlinear networks dense in a function space. The goodness-of-fit error decreases as the number of intermediate (hidden layer) outputs increases. For a 2-n-1 network, theory indicates that the approximation error decreases inversely proportional to $\sqrt{n}$ (i.e., quadrupling the number of hidden layer outputs should halve the error). Even though many of these theoretical results are not constructive, there exists powerful time-tested algorithms, such as backpropagation, that yield approximations that are consistent with theory. Implementations of back propagation are described in [11-13].

## 5 Results of Machine Learning on the Xor Approximation Problem

We showed in Section 1 that there is no linear formula that solves the xor representation or approximation problem. We showed in Section 3 that by extension, there is no linear feedforward network that solves the xor representation or approximation problem. However, Section 4 concluded by observing that nonlinear feedforward networks can approximate any bounded continuous function. In this Section we see how well backpropagation performs on the xor approximation problem. Note that backpropagation is a search procedure: an iterative form of the method used in calculus that finds critical points of the weights represented as an error surface. From a starting point of random weights, it picks an input (from the given set of input samples: also called the in-sample training set) and computes the corresponding output and slopes (gradients) with respect to the error surface. It uses the slopes to adjust the weights. After successive steps, we meander through the error surface over the in-sample data to find a bottom. The goal (or hope) is that the sequence of steps will get closer and closer to an error minimum.

Here are some examples. Figure 5.1 shows several $(x_1, x_2, z)$ charts with height $z = \text{out}(x_1, x_2)$ for the 2-2-1 topology with the random starting weights of Figure 4.1. We show several combinations of activation functions and use the 4-sample xor Boolean specification of Figure 1.1 as the in-sample training set. Depending on the activation function, these initial (random) candidate solutions are either smooth or have corners and folds.

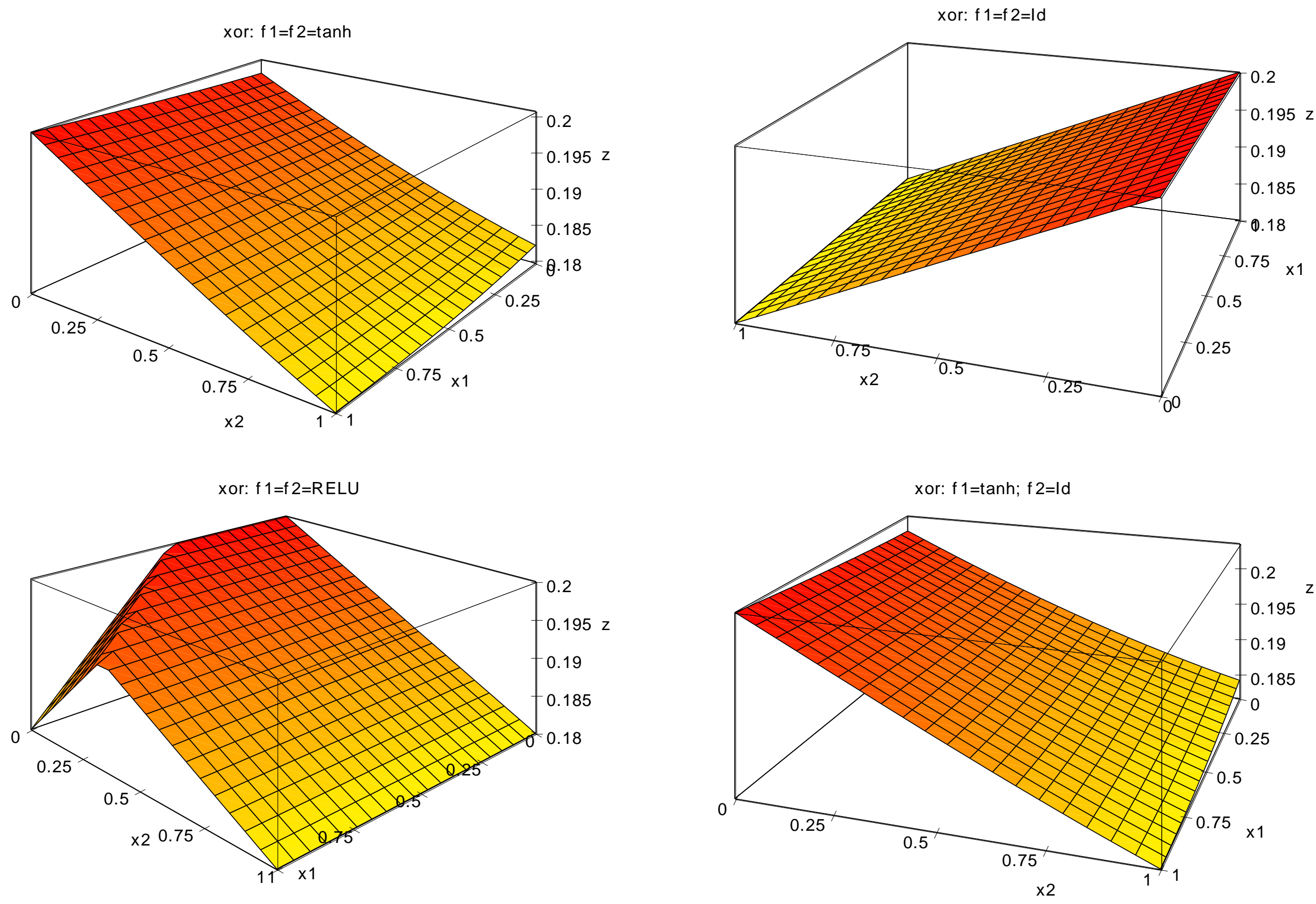

**Figure 5.1. Initial output functions using random weights from Figure 3.2 for the xor problem under 2-2-1 topology. Top left: inp-tanh-tanh;; top right:inpp-Id-Id (linear); bottom left: inp-RELU-RELU; bottom right: inp-tanh-Id.**

The figures show the results of backpropagation for the 2-2-1 network with different activation functions. The in-sample training set was the 4-sample Boolean xor specification (Figure 1.1). Training stopped after a few hundred iterations when the goodness-of-fit error was less than 0.001.

## 5.1 Xor with tanh-tanh Approximation

Figure 5.2 shows two solutions to the xor approximation problem, corresponding to two separate runs of backpropagation, for the 2-2-1 inp-tanh-tanh network. The algorithm derived two similar solutions (each solution corresponds to two different sets of weights); the resultant outputs for in-sample (the 4 Boolean inputs charted as the four corners in $(x_1, x_2)$ space) and out-sample (all points interior to the corners: $0 < x_1, x_2 < 1$) are shown on the charts below. The charts show the learned outputs as a $(x_1, x_2, z)$ chart with height $z = \text{out}(x_1, x_2)$ together with, for comparison, the optimal regression solution $\text{out}(x_1, x_2) = 0.5$.

| inp | w_1 | | | out_1 | w_2 | | (bias) | out | Note |
|---|---|---|---|---|---|---|---|---|---|
| 0.75 | *2.320* | *2.331* | -0.850 | 0.967746 | *2.68* | *-2.704* | *-0.826* | 0.994616 | ??? |
| 0.5 | *1.743* | *1.747* | -2.653 | -0.44002 | | | | | inp-tanh-tanh |
| 1 | | | | 1 | | | | | |
| inp | w_1 | | | out_1 | w_2 | | | out | Note |
| 0.75 | *1.665276* | *1.665982865* | -2.42363 | -0.32898 | *-2.46545* | *2.490741* | *-0.71314* | 0.98685 | ??? |
| 0.5 | *2.318576* | *2.329324919* | -0.84434 | 0.967984 | | | | | inp-tanh-tanh |
| 1 | | | | 1 | | | | | |

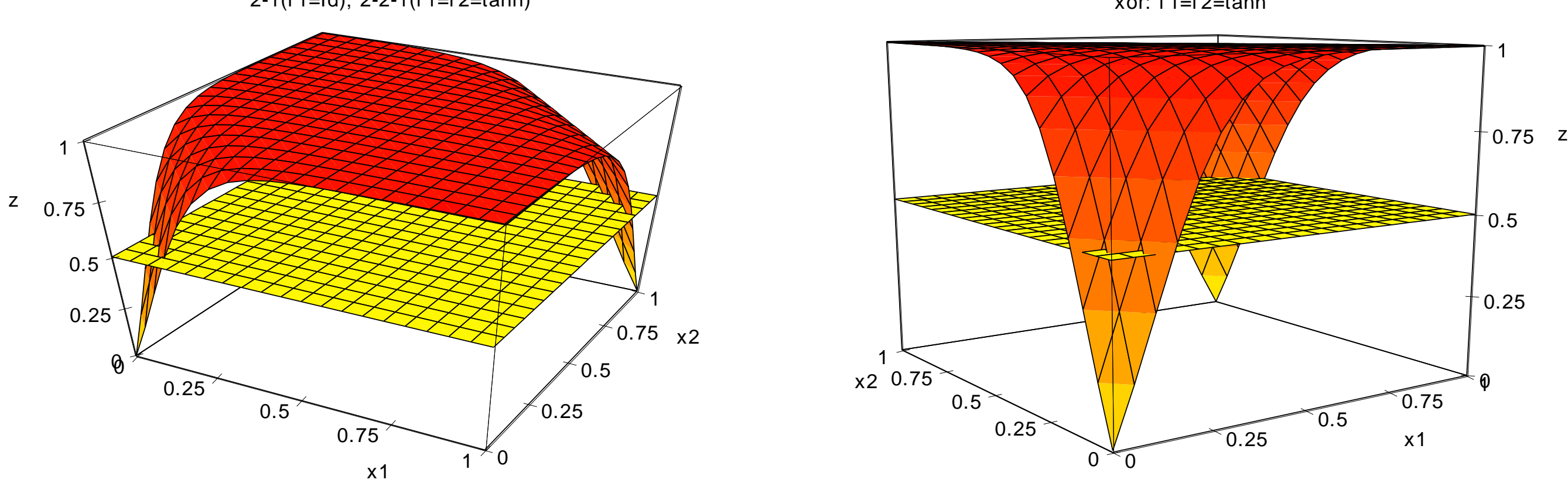

**Figure 5.2. Weights and output values for 2-2-1 topology with activation inp-tanh-tanh after machine learning via backpropagation.**

As expected, both outputs correctly interpolate and approximate xor to reasonable accuracy on the four input training samples: corners $(x_1, x_2) = (0,0), (0,1), (1,0), (1,1)$. All other non-corner $(x_1, x_2)$ points on the graph are out-sample values. The function corresponding to Figure 5.2 looks like it is learning – in the limit – a non-continuous function that is unity everywhere in $0 \le x_1, x_2 \le 1$ except at two corner endpoints when it is zero:

$$\text{out}(x_1, x_2) = u(|x_1 - x_2|) = \begin{cases} 0, & x_1 = x_2 = 0 \text{ or } x_1 = x_2 = 1; \\ 1, & \text{otherwise.} \end{cases}$$

## 5.2 Xor with RELU-RELU Approximation

Figure 5.3 shows two solutions for the 2-2-1 inp-RELU-RELU network. As above, the algorithm derived two solutions corresponding to two different sets of weights; the resultant outputs for in-sample (the four corners) and out-sample (interior to the corners) are shown on the charts below. The charts show the learned outputs as a $(x_1, x_2, z)$ chart with height $z = \text{out}(x_1, x_2)$ together with, for comparison, the optimal regression solution $\text{out}(x_1, x_2) = 0.5$.

| inp | w_1 | | | out_1 | w_2 | | | out | Note |
|---|---|---|---|---|---|---|---|---|---|
| 0.75 | *0.809* | *-0.809* | 0.809 | 1.01125 | *-1.236* | *1.64* | *1* | 0.250295 | F0 |
| 0.5 | *1.220* | *-1.220* | 0.000 | 0.305 | | | | | inp-RELU-RELU |
| 1 | | | | 1 | | | | | |
| inp | w_1 | | | out_1 | w_2 | | | out | Note |
| 0.75 | *0.858317* | *0.858316791* | 2.21E-17 | 1.072896 | *1.165071* | *-1.75039* | *7.75E-16* | 0.75 | Finf |
| 0.5 | *1.142601* | *1.142601152* | -1.1426 | 0.28565 | | | | | inp-RELU-RELU |
| 1 | | | | 1 | | | | | |

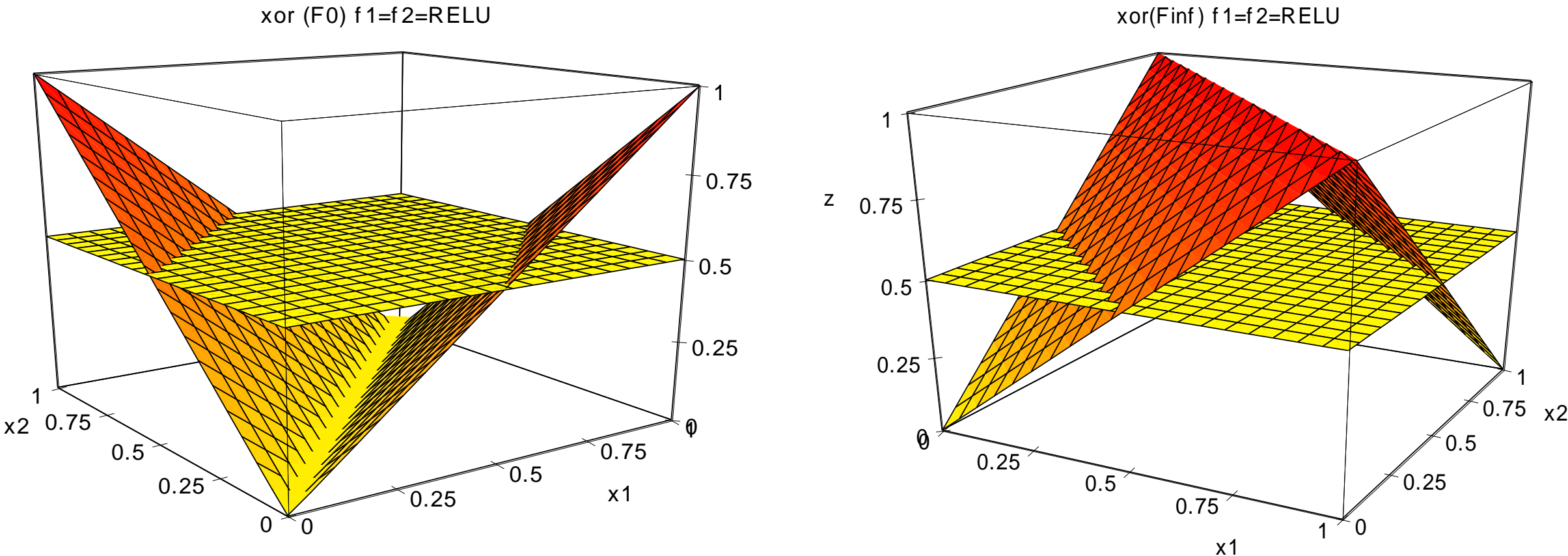

**Figure 5.3. Weights and output values for 2-2-1 topology with activation inp-RELU-RELU after machine learning via backpropagation.**

As expected, both outputs correctly interpolate and approximate xor to reasonable accuracy on the four input training samples: the corners $(x_1, x_2) = (0,0)$, $(0,1)$, $(1,0)$, $(1,1)$. All other non-corner $(x_1, x_2)$ points on the graph are out-sample values. The function corresponding to the first set of weights (charted in the bottom left of Figure 5.3) looks like Frank's copula-based continuous function with $s$=0. The function corresponding to the second set of weights (charted in the bottom right of Figure 5.3) also looks like Frank's copula-based continuous function with $s$=∞:

$$\mathrm{F}_0(x_1, x_2) = x_1 + x_2 - 2 \cdot \min(x_1, x_2) \text{ and } \mathrm{F}_\infty(x_1, x_2) = x_1 + x_2 - 2 \cdot \max(x_1 + x_2 - 1, 0).$$

This is a somewhat surprising result. Is this result to consistent with probabilistic logic? If so, we need to interpret all input and output values as probabilities. So for example, consider the input values shown in Figure 5.2 where $(x_1, x_2) = (0.75, 0.5)$. The output of machine learning for the first set of weights, is $F(0.75, 0.5) = 0.25$, consistent with $F_0$. The machine learning output for the second set of weights is $F(0.75, 0.5) = 0.75$, consistent with $F_\infty$. Clearly for these values the probabilistic inequality holds: $F_0(x_1,x_2) \le F_s(x_1,x_2) \le F_\infty(x_1,x_2)$. Both outputs are consistent with probabilistic logic and are reasonable approximations to the xor copula representations.

We repeated several experiments using different random starting points and different topologies. We were surprised to see that the majority of the converging inp-RELU-RELU and inp-RELU-ID networks converged to $\mathrm{F}_\infty\left(x_1,x_2\right)$; $\mathrm{F}_0\left(x_1,x_2\right)$ came in second. Figure 5.4 shows some other sample runs associated with the inp-RELU-RELU xor approximations.

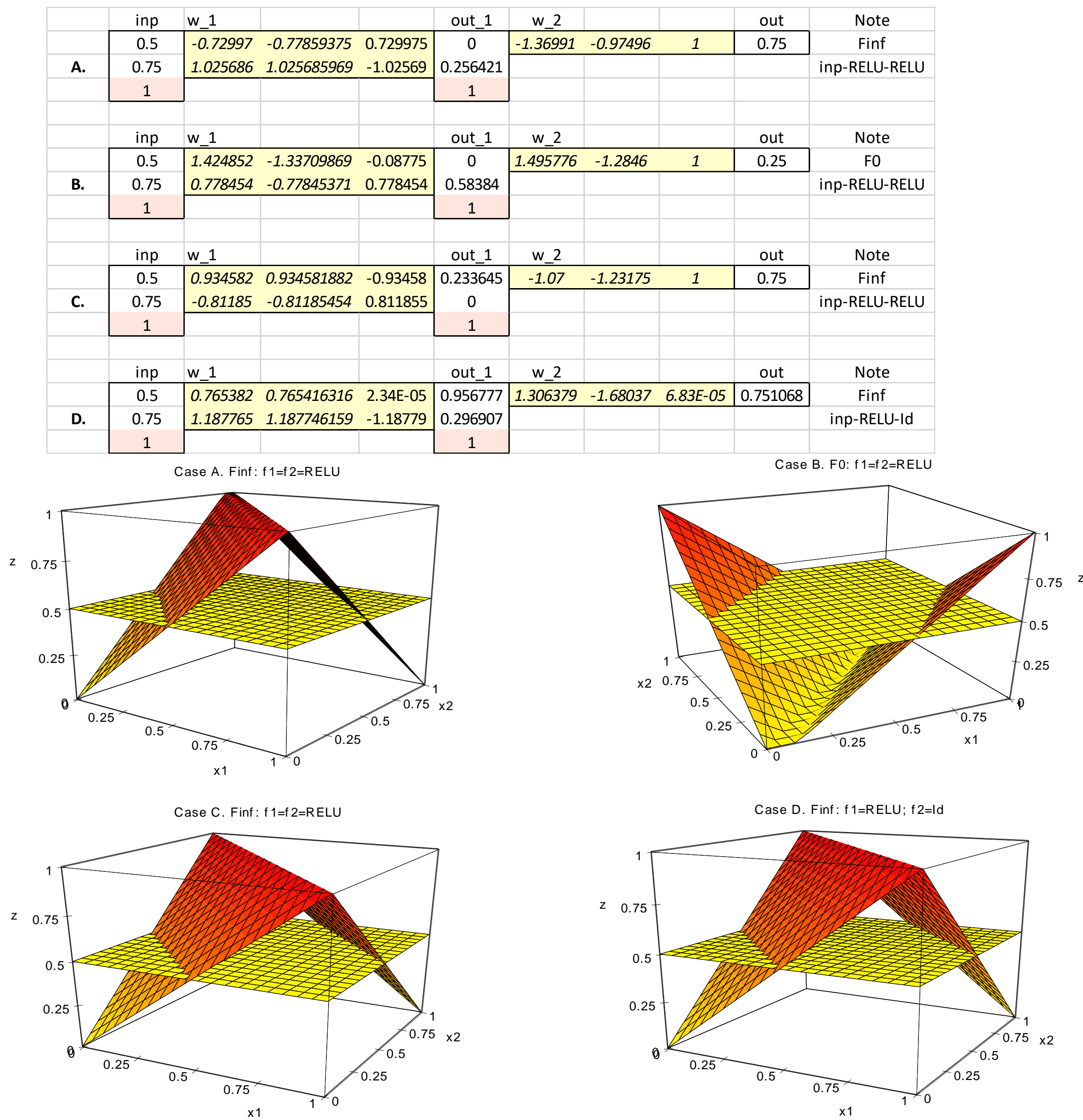

| | inp | w_1 | | | out_1 | w_2 | | | out | Note |
|---|---|---|---|---|---|---|---|---|---|---|
| | 0.5 | -0.72997 | -0.77859375 | 0.729975 | 0 | -1.36991 | -0.97496 | 1 | 0.75 | Finf |
| **A.** | 0.75 | 1.025686 | 1.025685969 | -1.02569 | 0.256421 | | | | | inp-RELU-RELU |
| | 1 | | | | 1 | | | | | |
| | | | | | | | | | | |
| | inp | w_1 | | | out_1 | w_2 | | | out | Note |
| | 0.5 | 1.424852 | -1.33709869 | -0.08775 | 0 | 1.495776 | -1.2846 | 1 | 0.25 | F0 |
| **B.** | 0.75 | 0.778454 | -0.77845371 | 0.778454 | 0.58384 | | | | | inp-RELU-RELU |
| | 1 | | | | 1 | | | | | |
| | | | | | | | | | | |
| | inp | w_1 | | | out_1 | w_2 | | | out | Note |
| | 0.5 | 0.934582 | 0.934581882 | -0.93458 | 0.233645 | -1.07 | -1.23175 | 1 | 0.75 | Finf |
| **C.** | 0.75 | -0.81185 | -0.81185454 | 0.811855 | 0 | | | | | inp-RELU-RELU |
| | 1 | | | | 1 | | | | | |
| | | | | | | | | | | |
| | inp | w_1 | | | out_1 | w_2 | | | out | Note |
| | 0.5 | 0.765382 | 0.765416316 | 2.34E-05 | 0.956777 | 1.306379 | -1.68037 | 6.83E-05 | 0.751068 | Finf |
| **D.** | 0.75 | 1.187765 | 1.187746159 | -1.18779 | 0.296907 | | | | | inp-RELU-Id |
| | 1 | | | | 1 | | | | | |

**Figure 5.4. Weights and output values for 2-2-1 topology with activation inp-RELU-RELU after machine learning via backpropagation. Top: weights shown as Cases A-D. Bottom: Charts corresponding to Cases A-D.**

## 6 Error Tabulation and Error Surfaces

Understanding convergence requires an understanding of the error surface. Recall the 2-1 linear network (used in linear regression) with output

$$\text{out} := w_{11} \cdot x_1 + w_{12} \cdot x_2 + w_{13}$$

The linear regression problem is to find values for the three weights so that the errors are as small as possible. Recall that we are able to visualize the output by plotting $(x_1, x_2, z)$ with height $z = \text{out}(x_1, x_2)$. For the output chart there are 2 degrees of freedom (dimensions) for the inputs: this implies that visualizing the problem requires charting a 2-dimensional surface in 3-dimensional space.

Using the 4-sample Boolean specification as the training set (Figure 1.1), the equation for the goodness-of-fit error (using sum of squares) is:

$$\begin{aligned} \text{err} &:= \left(\text{out}(0,0)-0\right)^2 + \left(\text{out}(0,1)-1\right)^2 + \left(\text{out}(1,0)-1\right)^2 + \left(\text{out}(1,1)-0\right)^2 \\ &= \left(w_3\right)^2 + \left(w_1 + w_2 + w_3\right)^2 + \left(w_1 - 1 + w_3\right)^2 + \left(-1 + w_2 + w_3\right)^2 \end{aligned}$$

Can we do the same thing with by plotting this goodness-of-fit error function? There are 3 degrees of freedom for the weights $(w_1, w_2, w_3)$: visualizing the problem requires plotting $(w_1, w_2, w_3, z)$ with height $z = \text{err}(w_1, w_2, w_3)$. This requires charting a 3-dimensional error surface in 4-dimensional space. Unfortunately our visual perception is limited to three dimensions.

The situation is more complex for nonlinear network topologies. The 2-2-1 nonlinear output is (Section 4)

$$\text{out} := f_2\left(w_{11}^2 \cdot \left(f_1\left(w_{11}^1 \cdot x_1 + w_{12}^1 \cdot x_2 + w_{13}^1\right)\right) + w_{12}^2 \cdot \left(f_1\left(w_{21}^1 \cdot x_1 + w_{22}^1 \cdot x_2 + w_{23}^1\right)\right) + w_{13}^2\right)$$

For the 4-sample Boolean training set, the equation for the goodness-of-fit is:

$$\begin{aligned} \text{err} :=& \left(f_2\left(w_{13}^2 - w_{11}^2 \cdot f_1(w_{13}^1) + w_{12}^2 \cdot f_1(w_{23}^1)\right)\right)^2 \\ &+ \left(f_2\left(w_{13}^2 + w_{11}^2 \cdot f_1(w_{11}^1 + w_{12}^1 + w_{13}^1) + w_{12}^2 \cdot f_1(w_{21}^1 + w_{22}^1 + w_{23}^1)\right)\right)^2 \\ &+ \left(f_2\left(w_{13}^2 + w_{11}^2 \cdot f_1(w_{11}^1 + w_{13}^1) + w_{12}^2 \cdot f_1(w_{21}^1 + w_{23}^1)\right) - 1\right)^2 \\ &+ \left(f_2\left(w_{13}^2 + w_{11}^2 \cdot f_1(w_{12}^1 + w_{13}^1) + w_{12}^2 \cdot f_1(w_{22}^1 + w_{23}^1)\right) - 1\right)^2 \end{aligned}$$

Visualizing the problem requires plotting $\left(w_{11}^1, w_{12}^1, w_{13}^1, w_{21}^1, w_{22}^1, w_{23}^1, w_{11}^2, w_{12}^2, w_{13}^2, z\right)$ with height $z = \text{err}\left(w_{11}^1, w_{12}^1, w_{13}^1, w_{21}^1, w_{22}^1, w_{23}^1, w_{11}^2, w_{12}^2, w_{13}^2\right)$. For the goodness-of-fit plot there are 9 degrees of freedom (dimensions) for the weights: visualizing the problem requires charting a 9-dimensional error surface in 10-dimensional space. One solution is to restrict ourselves to 2-dimensional error surface slices. For example, let's simplify the problem to 2 degrees of freedom by assigning the values tabulated in Figure 3.2 to all weights except for $w_{11}^1$ and $w_{12}^1$. This error surface formula – called the $w_{11}^1 \times w_{12}^1$ projection – is:

$$\begin{aligned} \text{err}(w_{11}, w_{12}) := &\left( f_2\left(0.3 - 0.4 \cdot f_1(0.2) - 0.2 \cdot f_1(0.1)\right)\right)^2 \\ &+ \left( f_2\left(0.3 - 0.4 \cdot f_1(w_{11}^1 + w_{12}^1 + 0.2) - 0.2 \cdot f_1(0.2)\right)\right)^2 \\ &+ \left( f_2\left(0.3 - 0.4 \cdot f_1(w_{11}^1 + 0.2) - 0.2 \cdot f_1(-0.1)\right) - 1\right)^2 \\ &+ \left( f_2\left(0.3 - 0.4 \cdot f_1(w_{12}^1 + 0.2) - 0.2 \cdot f_1(0.4)\right) - 1\right)^2 \end{aligned}$$

There is nothing special about $w_{11}^1$ and $w_{12}^1$. We could have selected any two weights to get a 2-dimensional projection of the 9-dimensional surface. These are straightforward to configure and plot with any computer algebra system.

The following pages show several three-dimensional charts for different projections of the 2-dimensional error surface for the xor problem. For the 2-2-1 topology, there are

$$\binom{9}{2} = 36$$

projections. The horizontal (xy) plane shows the values in the two-dimensional weight-space; the vertical (z axis) shows the value of the error surface (sums of squares of the errors over all four samples). We also show these values represented as a three-dimensional contour display. Our parameters are the different output functions:

$$\text{Id}(t) = t;\ \tanh(t); \text{RELU}(t)$$

The first figure shows the case where $f_1(z) = f_2(z) = z$: this is the same case as the linear network family in Section 3. The equation for the error surface using sum of squared errors is:

$$\begin{aligned} \text{err}(w_{11}, w_{12}) := &\left(0.18 - 0.4 \cdot w_{11} - 0.4 \cdot w_{12}\right)^2 - \left(0.04 \cdot w_{12} + 0.06\right) \cdot x_2 + 0.2 \\ &+ \left(-0.4 \cdot w_{11} - 0.76\right)^2 + \left(-0.4 \cdot w_{12} - 0.86\right)^2 + 0.04 \end{aligned}$$

As expected, the error surface is a paraboloid bowl with a single global minimum (a unique solution) and no valleys or hidden peaks. The other nonlinear functions show the complexity of the error surfaces.

Corners and folds present problems for gradient search methods: these shapes induce non-continuous slopes. Moreover, nonlinear surfaces are might lead candidate solutions to get suck in local valleys or on non-optimal local peaks or "saddle points." We can see examples of these with the error surface charts in Figures 6.1-6.7.

**FIGURE 6.1. Identity function. Linear Family. For 2-2-1: inp-Id-Id.**

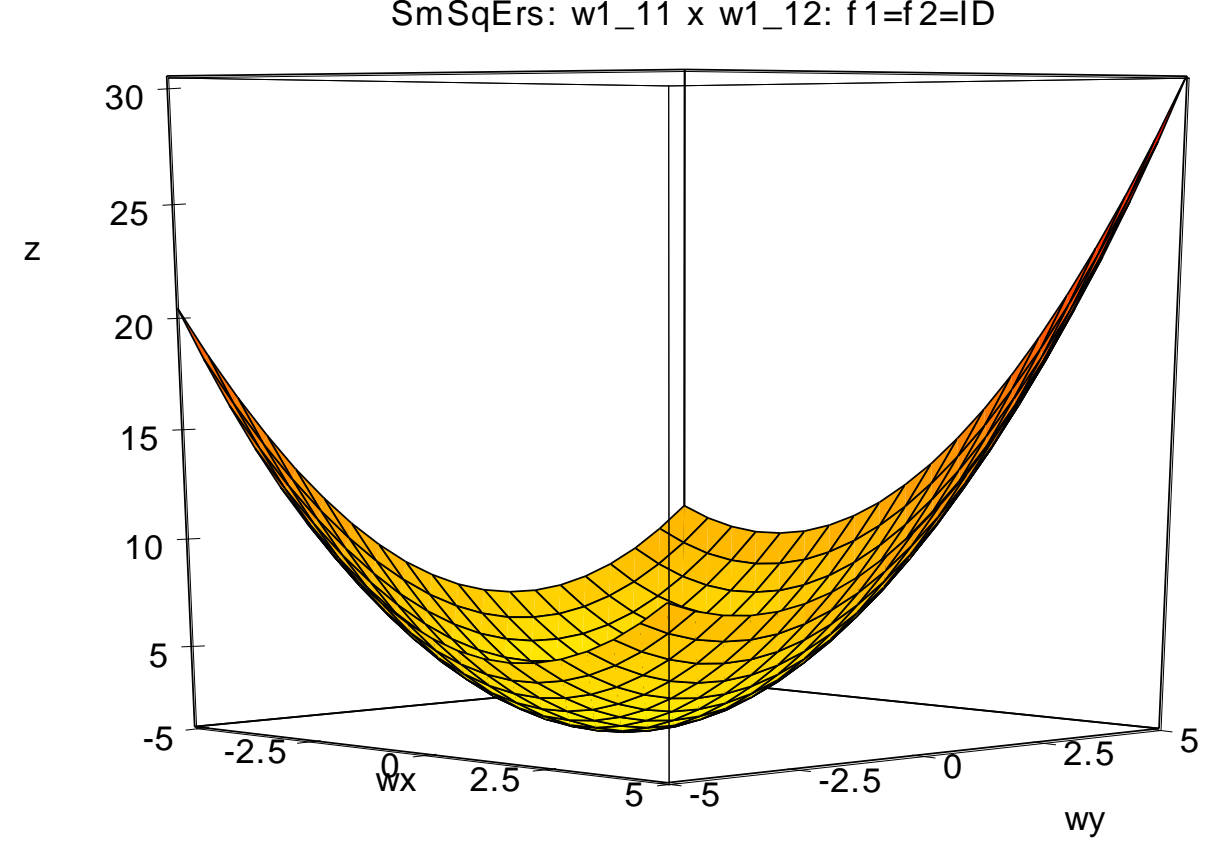

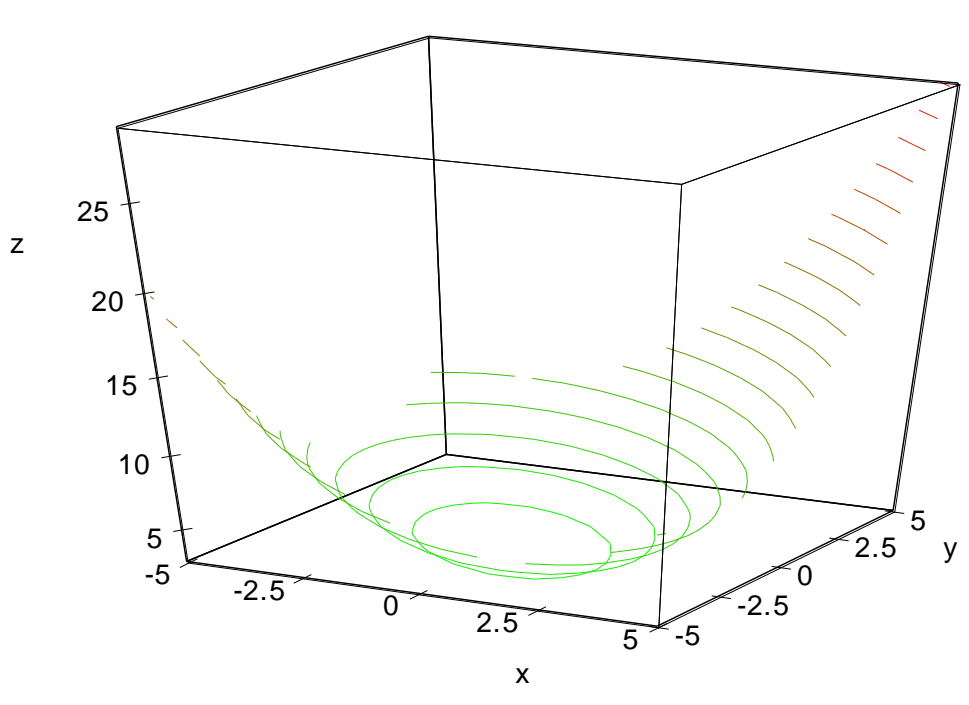

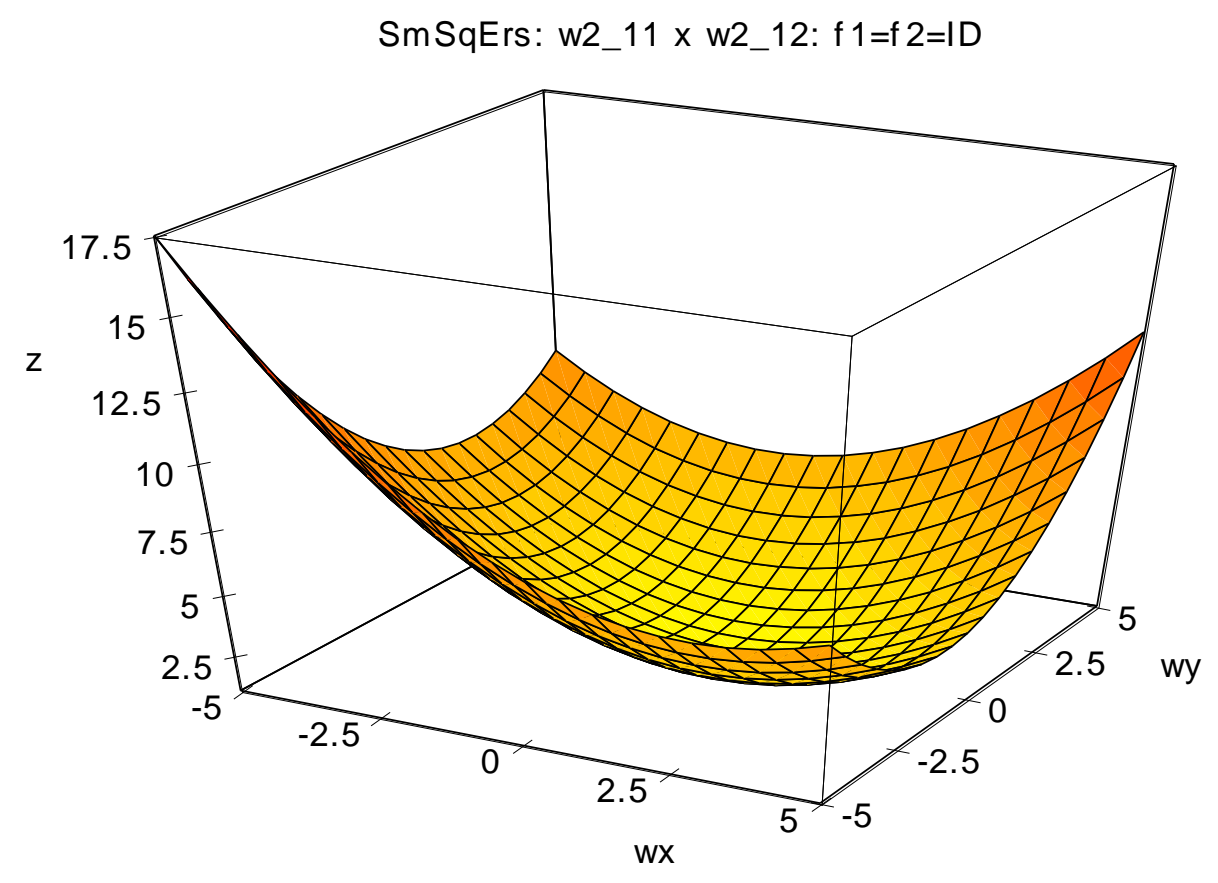

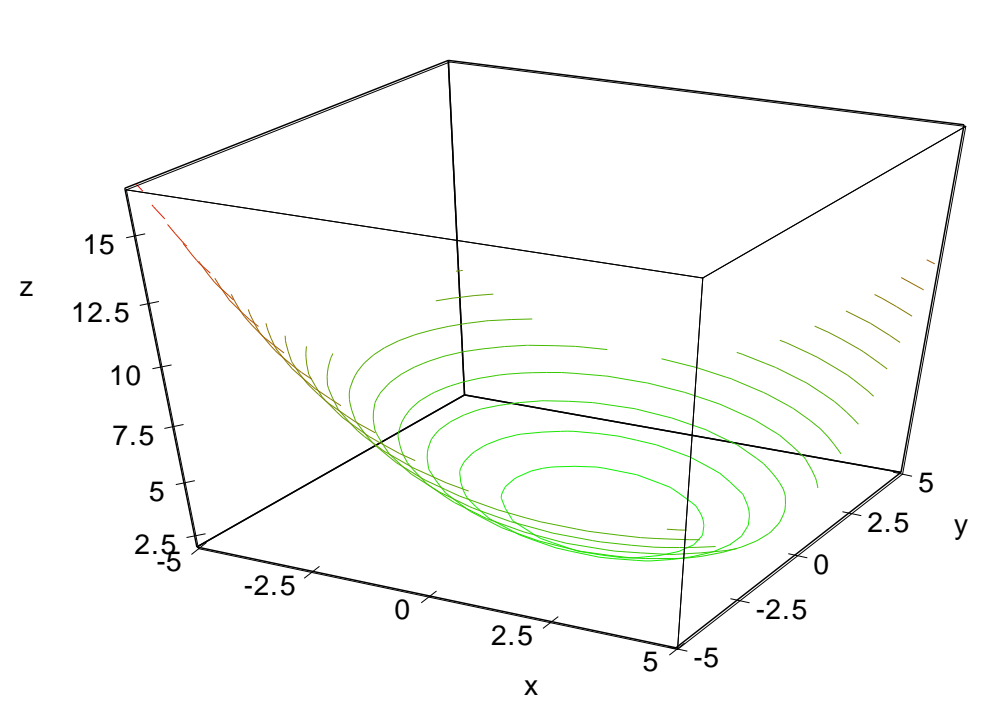

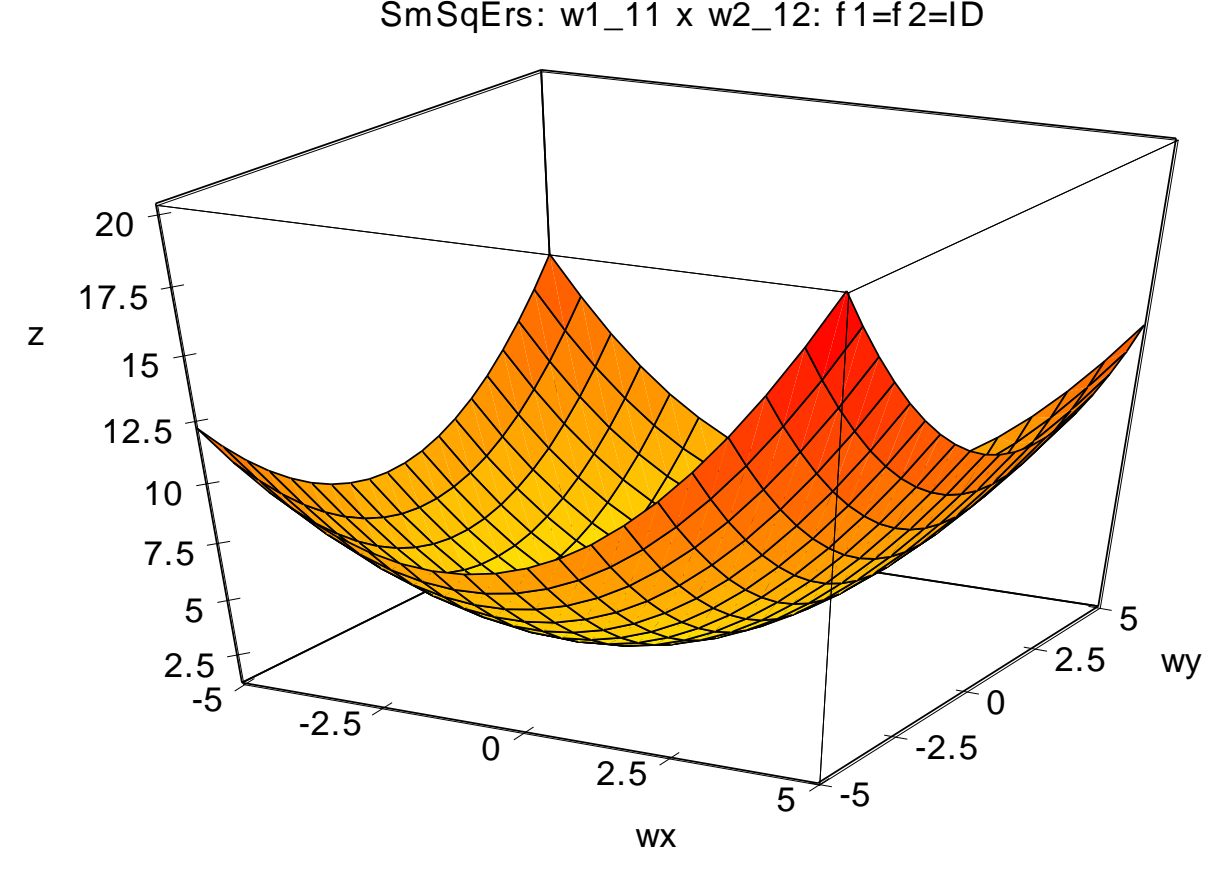

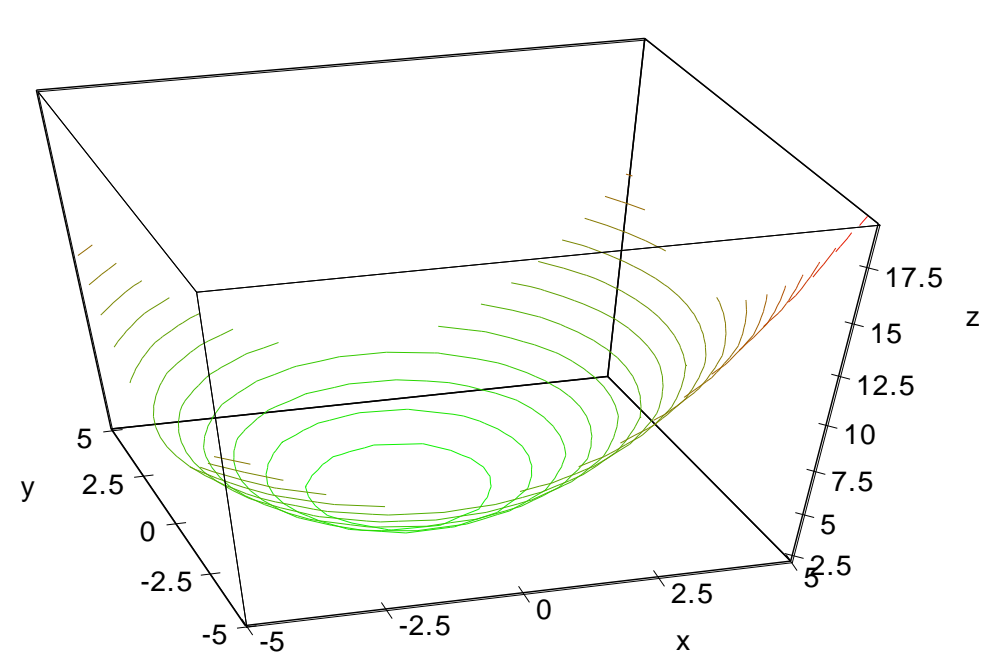

**FIGURE 6.2. RELU function. RELU Family. For 2-2-1: inp-RELU-RELU:**

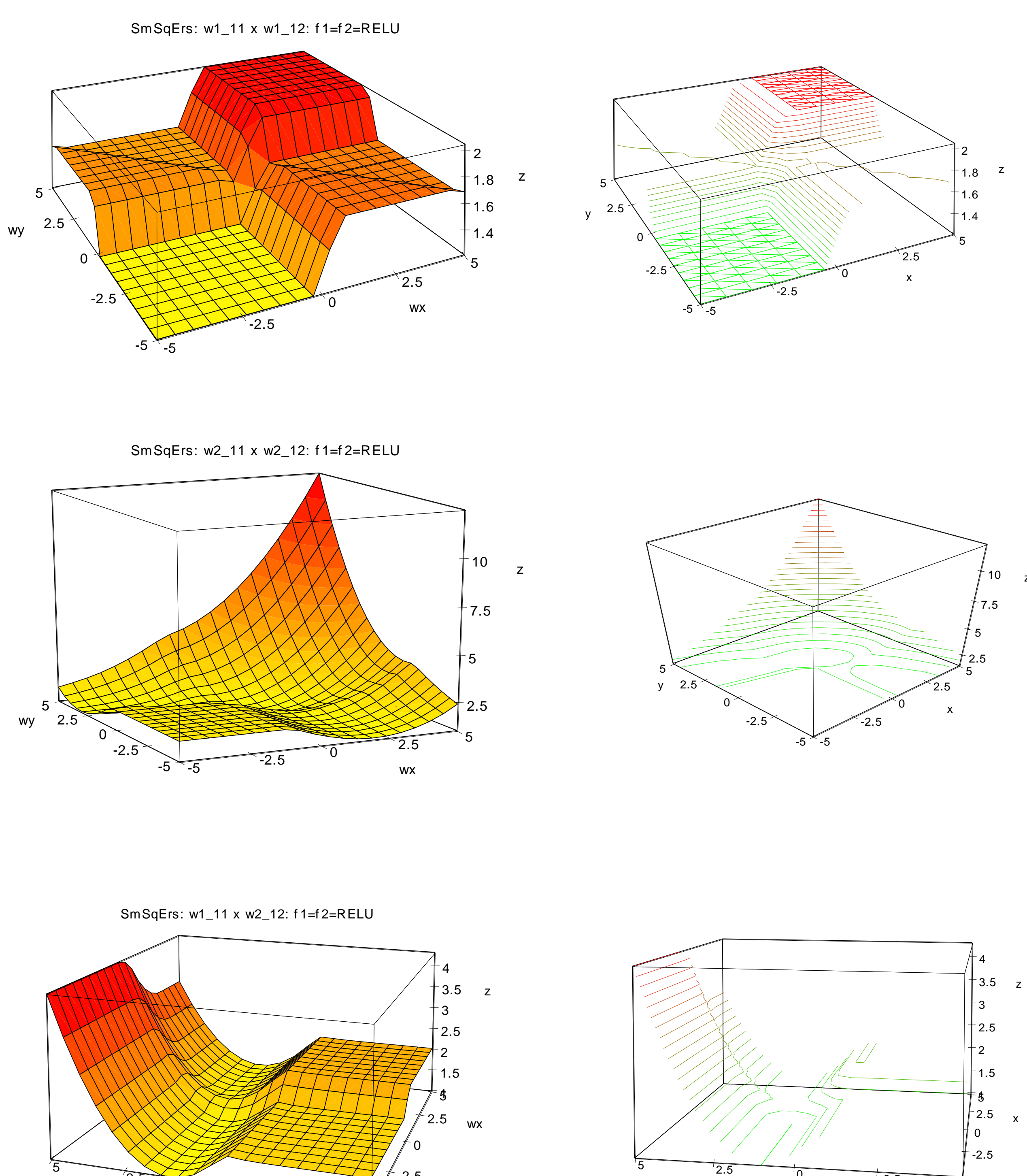

**FIGURE 6.3. Ridge Functions. For 2-2-1:. Surface w1_11 x w1_12.**

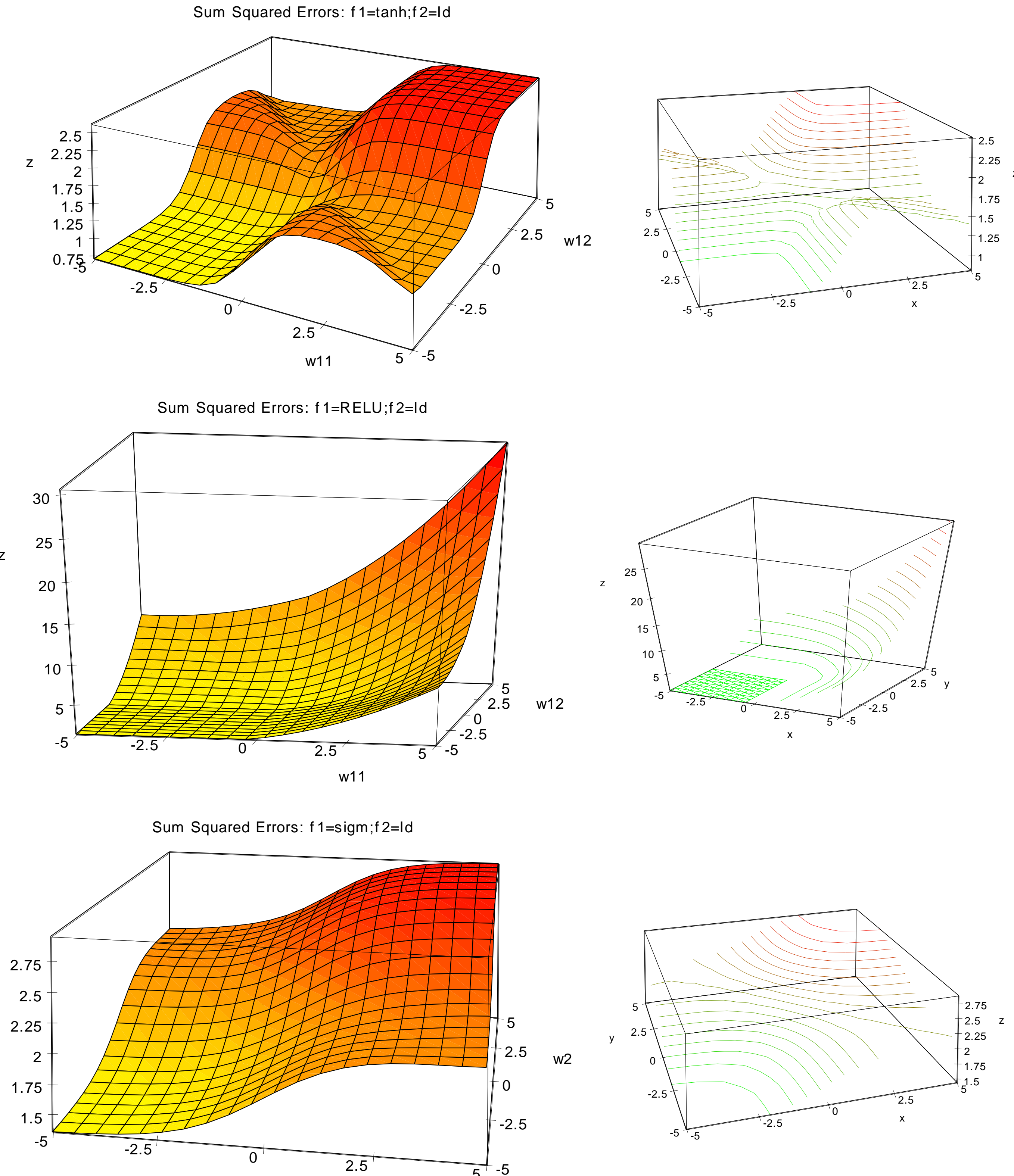

**FIGURE 6.4. Same activation function for both layers: f1=f2. Surface w1_11 x w1_12.**

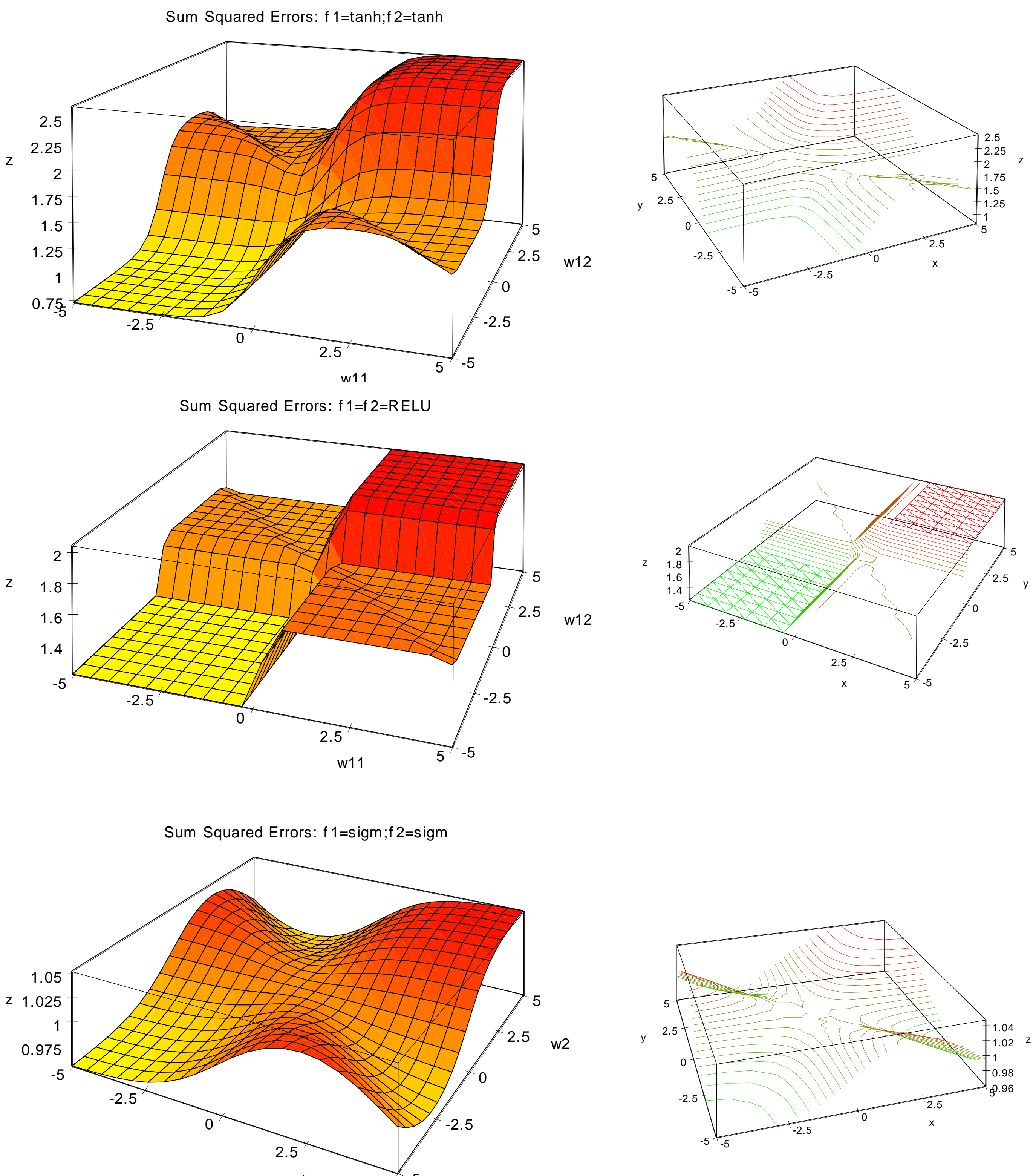

**FIGURE 6.5. Identity function for first layer: f1=Id. Surface w1_11 x w1_12.**

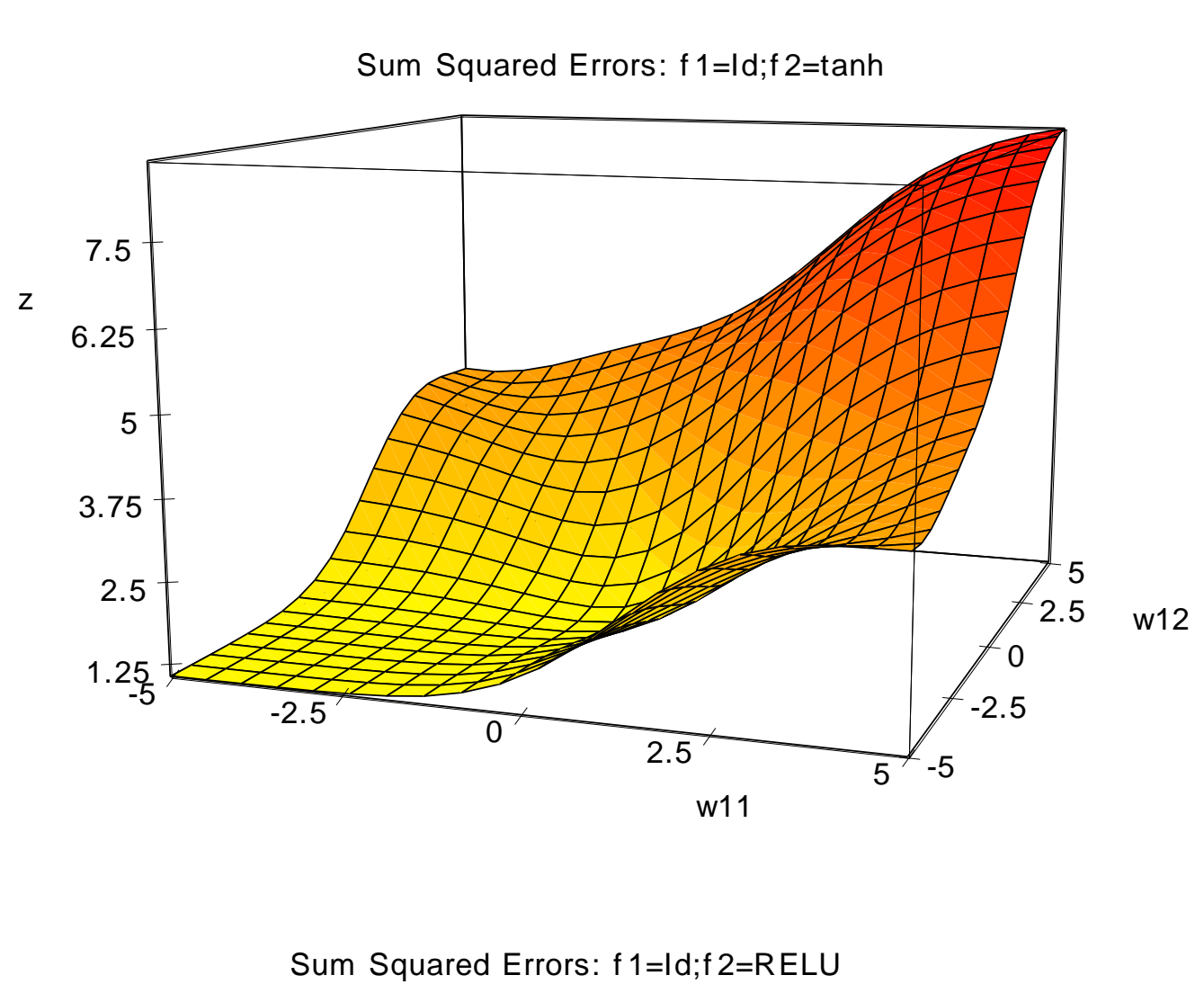

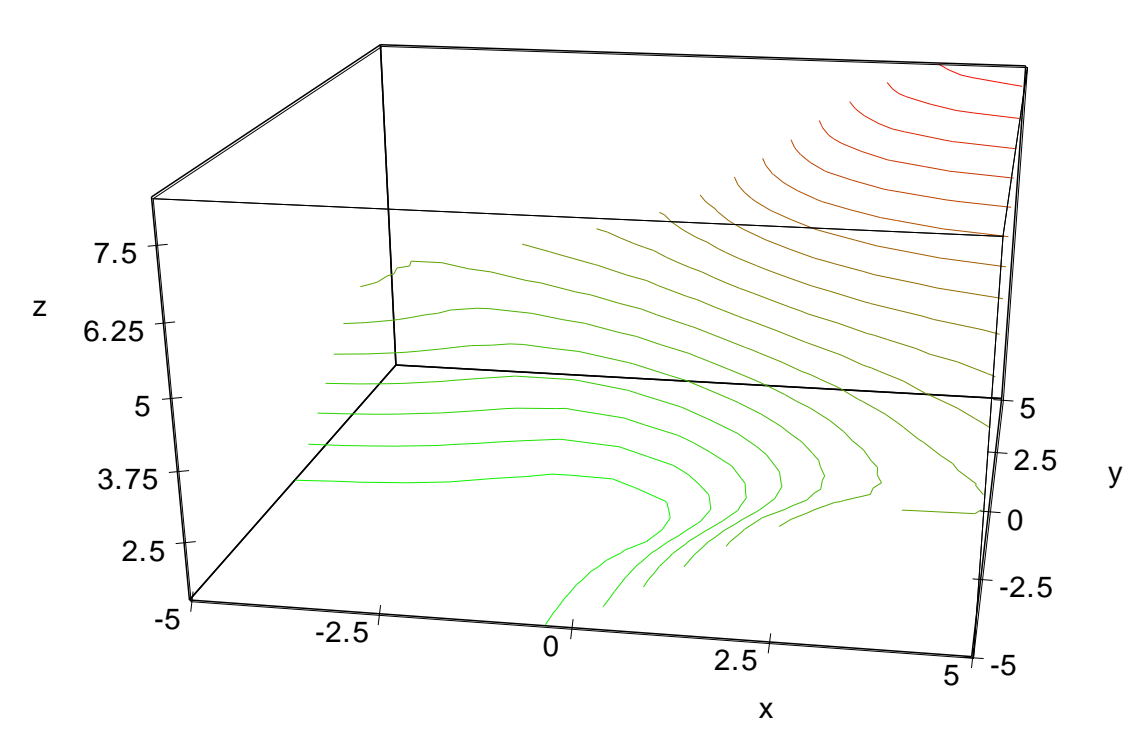

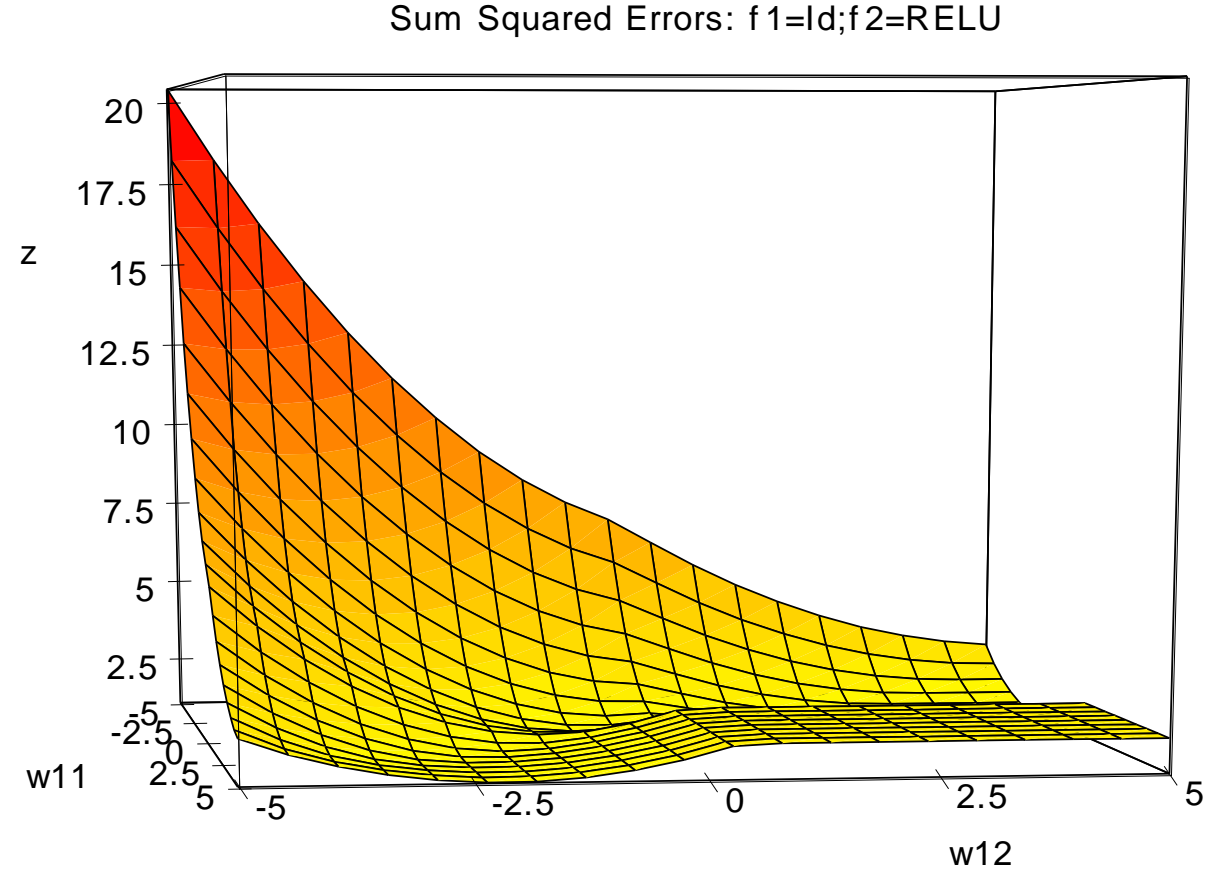

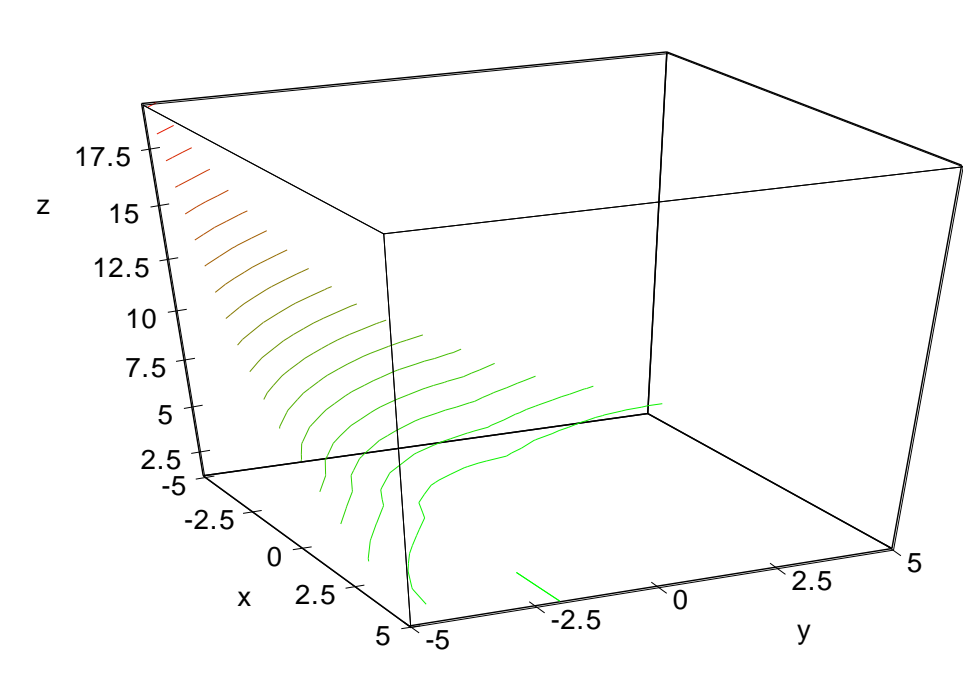

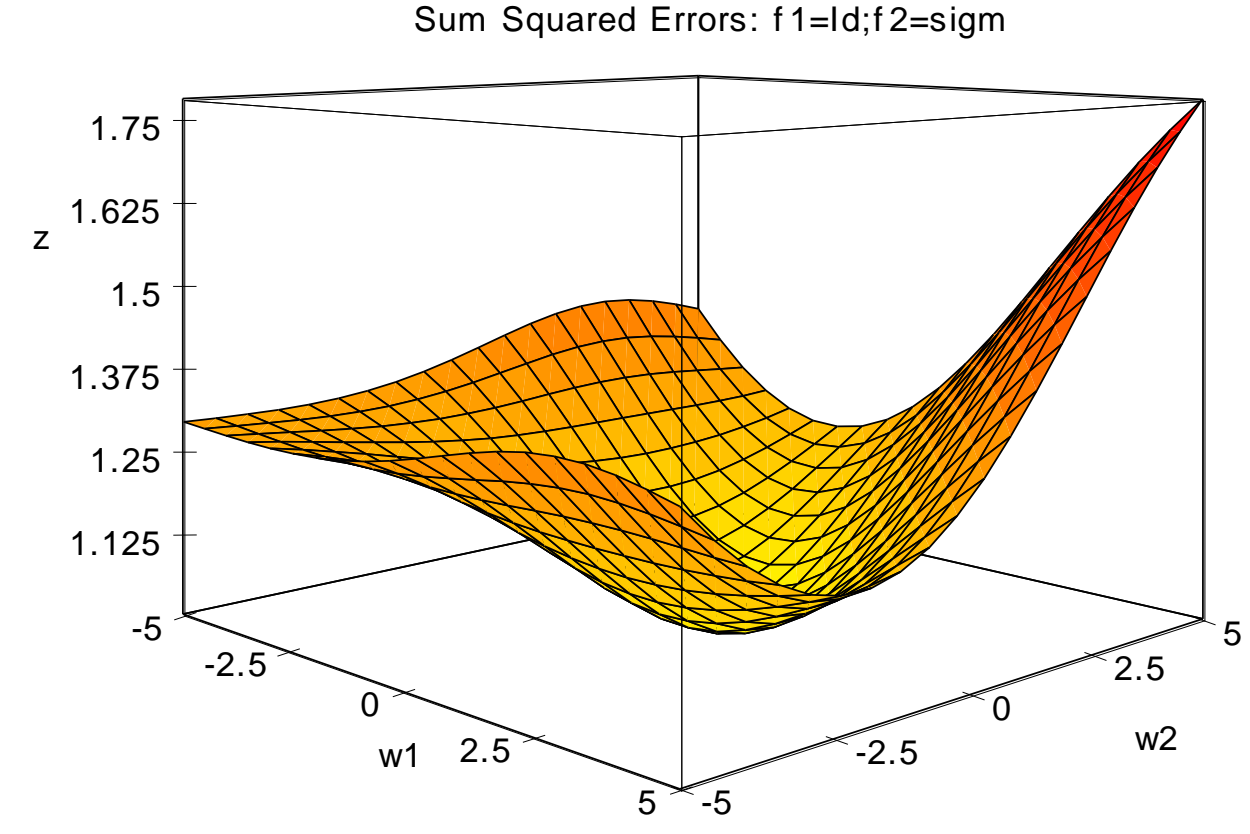

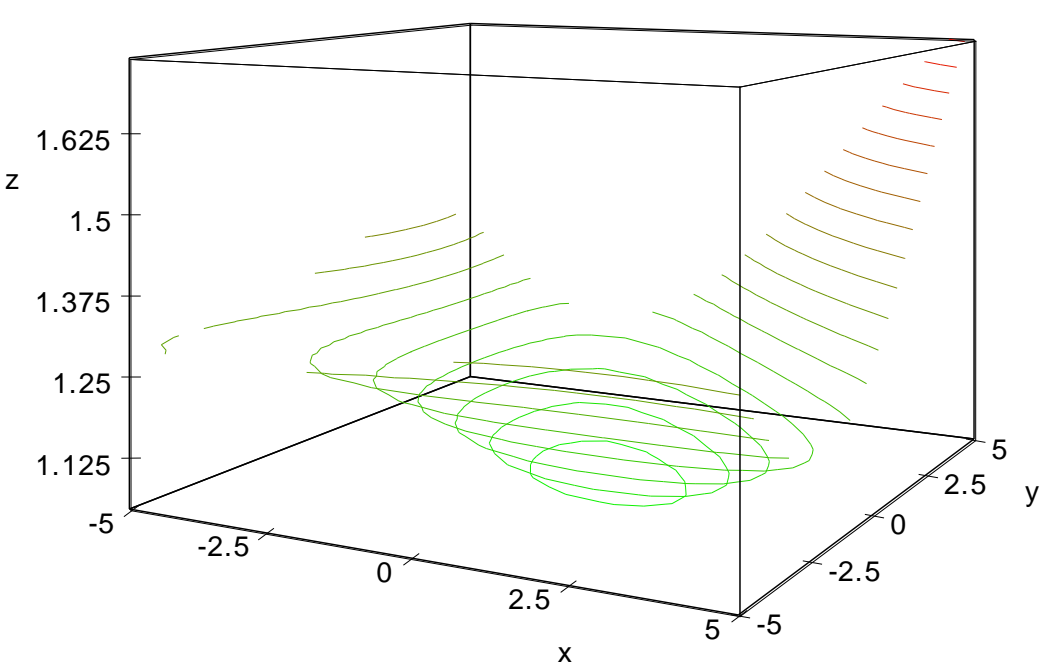

**FIGURE 6.6. Same activation function for both layers: f1=f2. Surface w1_11 x w1_21.**

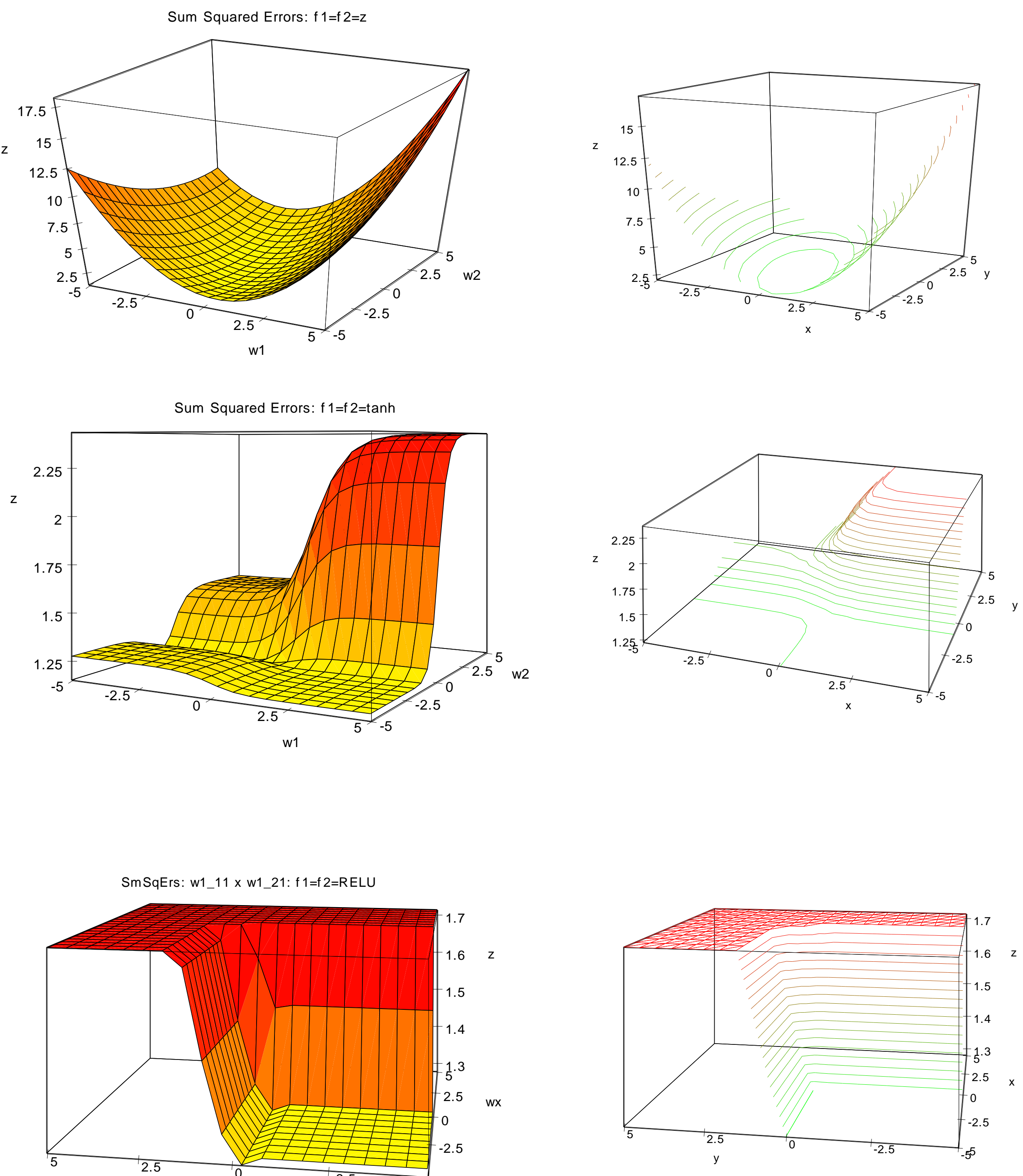

**FIGURE 6.7. Same activation function for both layers: f1=f2. Surface w1_11 x w2_12.**

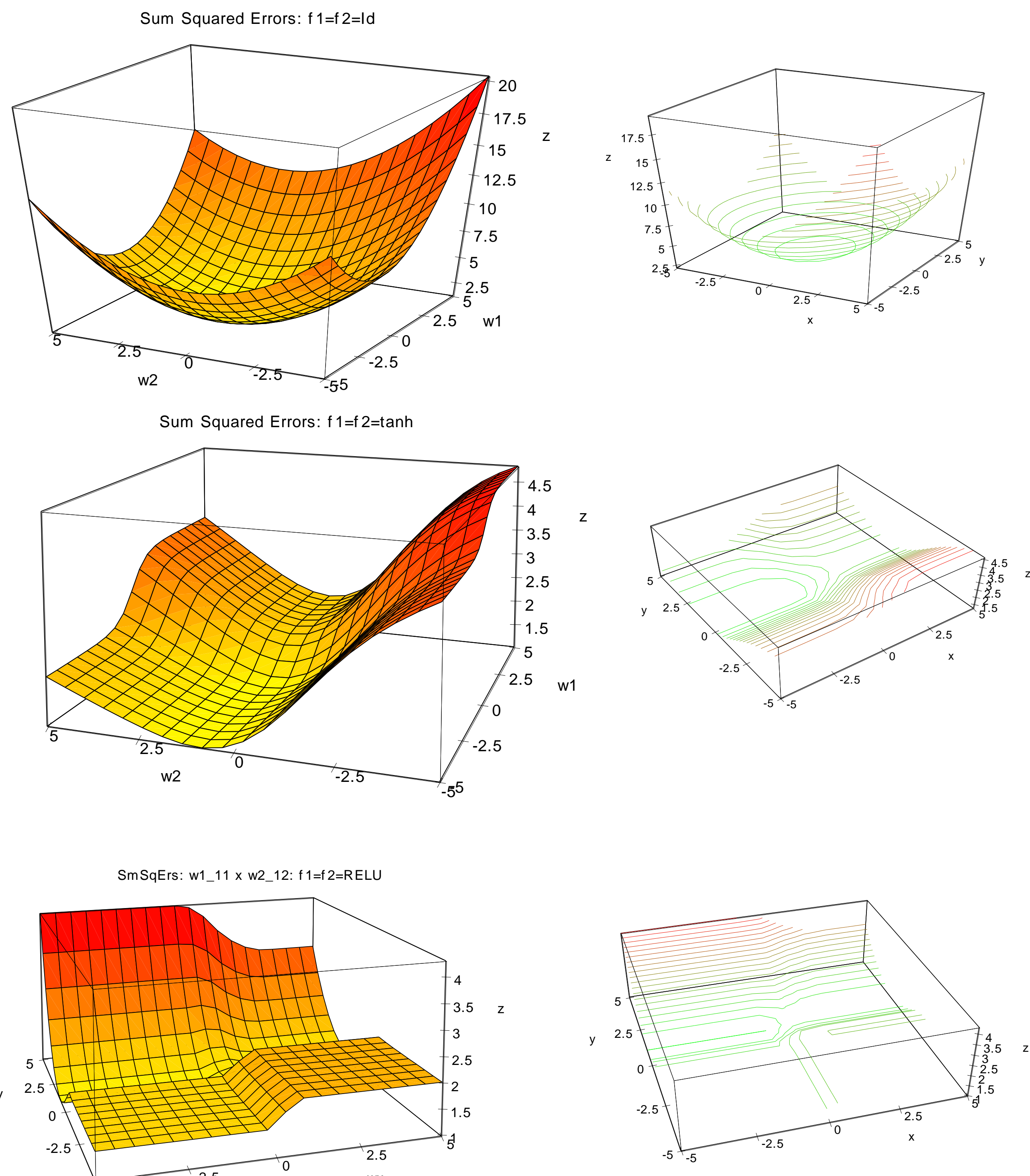

## 7. Discussion

The 2-2-1 xor approximation problem using 4 Boolean inputs as the training set should be simple enough to completely analyze. Part of our motivation was to try to understand why the simple 2-2-1 inp-RELU-RELU networks trained to solve the xor representation problem were able to generalize its results to copula-based xor representations on an out-sample data set of real values between zero and one. Comparing the outputs of a machine learning experiment with values not seen in the training set is called cross-validation. In the following, we examine the copula representations in the context of selecting in-sample and out-sample data for cross-validation.

### 7.1 From Boolean Values to Analog Values

In Section 2, we have seen that, in the context of probabilistic logic, copulas provide a consistent way to extend a set of Boolean values to a set of analog values for both inputs and outputs. If a logical statement can be represented in terms of "and", "or," and "not", then the probability of the statement can be represented in terms of copulas $A_s(x, y)$ and $R_s(x, y)$.

The charts shown in Section 5 showed the 2-2-1 networks trained on 4 Boolean inputs: the corners $(x_1, x_2) = (0,0), (0,1), (1,0), (1,1)$. All other non-corner $(x_1, x_2)$ points on the charts are out-sample values. Comparing the outputs of a machine learning experiment with values not seen in the training set is called cross-validation. Let's examine how the copula representations helps in selecting in-sample and out-sample data for cross-validation. Let $X_1$ and $X_2$ be statements: logical sentences that are either TRUE (1) or FALSE (0). Let $x_1 = \Pr\left[X_1\right]$ and $x_2 = \Pr\left[X_2\right]$. The plausibility for xor, $F\left(x_1, x_2\right) = \Pr\left[X_1 \text{ xor } X_2\right]$, can be presented in the following ways:

Boolean specification:
$$\begin{cases} F\left(1,0\right) = F\left(0,1\right) = 1 \\ F\left(0,0\right) = F\left(1,1\right) = 0 \end{cases}$$

Analog specification:
$$\begin{cases} F\left(1,x\right) = F\left(x,1\right) = 1 - x \\ F\left(0,x\right) = F\left(x,0\right) = x \end{cases}$$

Copula specification:
$$\mathrm{F}\left(x_1, x_2\right) = \begin{cases} \mathrm{F}_0\left(x_1, x_2\right) = x_1 + x_2 - 2 \cdot \min(x_1, x_2) \\ \mathrm{F}_s\left(x_1, x_2\right) = x_1 + x_2 - 2 \cdot \log_s\left(1 + \dfrac{\left(s^{1-x_1} - 1\right) \cdot \left(s^{1-x_2} - 1\right)}{s - 1}\right) \\ \mathrm{F}_1\left(x_1, x_2\right) = x_1 + x_2 - 2 \cdot x_1 \cdot x_2 \\ \mathrm{F}_\infty\left(x_1, x_2\right) = x_1 + x_2 - 2 \cdot \max(x_1 + x_2 - 1, 0) \end{cases}$$

Each specification specifies a training set for machine learning or for cross-validation. For example, the 2-2-1 networks in Section 5 were all trained with the 4-sample Boolean Specification. What about cross-validation? We showed that many runs of the inp-RELU-RELU network converged to the copula-based representations $\mathrm{F}_0$ or $\mathrm{F}_\infty$, and the inp-tanh-tanh network converged to another function not consistent with probabilistic logic. This shows that one in-sample specification can result in many valid trained networks with very different characteristics.

Note that the xor copula representations for $\mathrm{F}_0$ and $\mathrm{F}_\infty$ have the following RELU representations:

$$\mathrm{F}_0\left(x_1,x_2\right)=x_1+x_2-2\cdot\min(x_1,x_2)=\mathrm{RELU}\left(\mathrm{RELU}\left(x_1-x_2\right)+\mathrm{RELU}\left(x_2-x_1\right)\right)$$
$$\mathrm{F}_\infty\left(x_1,x_2\right)=x_1+x_2-2\cdot\max(x_1+x_2-1,0)=\mathrm{RELU}\left(\mathrm{RELU}(x_1+x_2)-2\cdot\mathrm{RELU}(x_1+x_2-1)\right)$$

These representations are consistent with the 2-2-1 approximations where (see Section 4):

$$\mathrm{out}=\mathrm{RELU}\left(w_{11}^2\cdot\left(\mathrm{RELU}\left(w_{11}^1\cdot x_1+w_{12}^1\cdot x_2+w_{13}^1\right)\right)+w_{12}^2\cdot\left(\mathrm{RELU}\left(w_{21}^1\cdot x_1+w_{22}^1\cdot x_2+w_{23}^1\right)\right)+w_{13}^2\right)$$

These copula-based representations for $\mathrm{F}_0$ and $\mathrm{F}_\infty$ are shown in Figure 7.1.

| inp | w_1 | | | out_1 | w_2 | | | out | Note |
|---|---|---|---|---|---|---|---|---|---|
| 0.75 | 1 | -1 | 0 | 0.25 | 1 | 1 | 0 | 0.25 | F0 |
| 0.5 | -1 | 1 | 0 | 0 | | | | | inp-RELU-RELU |
| 1 | | | | 1 | | | | | |
| | | | | | | | | | |
| inp | w_1 | | | out_1 | w_2 | | | out | Note |
| 0.75 | 1 | 1 | 0 | 1.25 | 1 | -2 | 0 | 0.75 | Finf |
| 0.5 | 1 | 1 | -1 | 0.25 | | | | | inp-RELU-RELU |
| 1 | | | | 1 | | | | | |

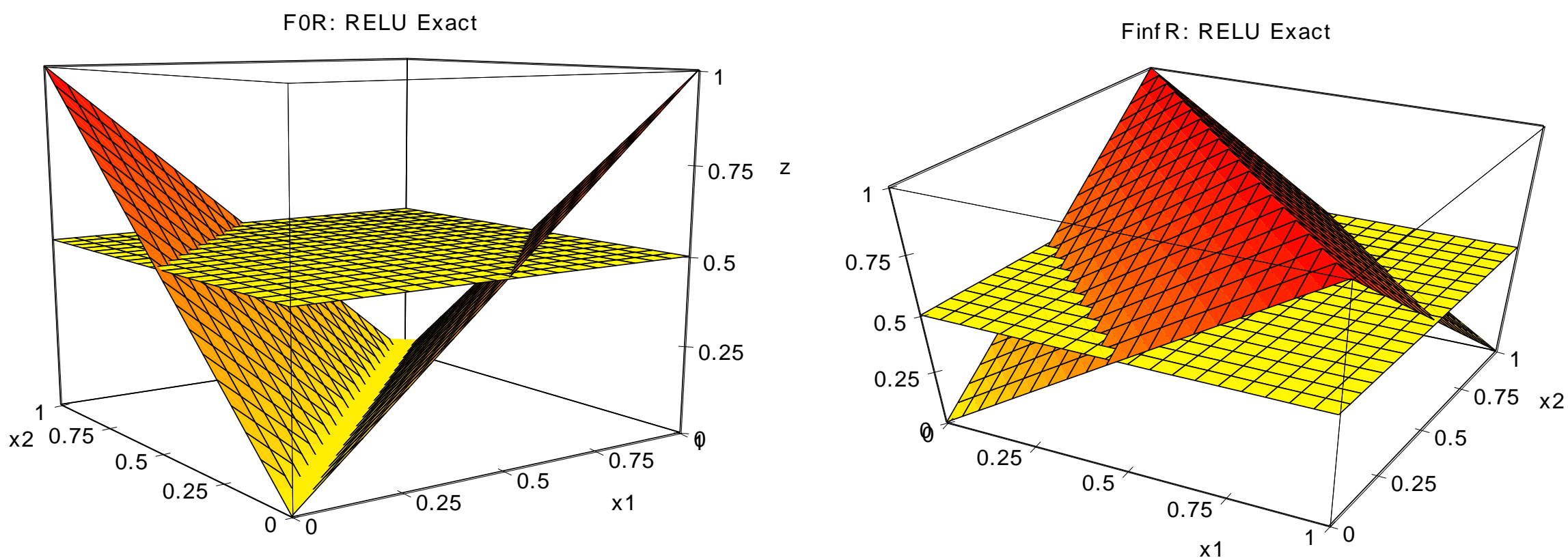

**Figure 7.1. Weights and output values for 2-2-1 topology with activation inp-RELU-RELU for exact network representations for $\mathrm{F}_0$ and $\mathrm{F}_\infty$, together with linear regression. Compare Figures 5.3 and 5.4.**

Perhaps for Boolean functions based on logical operations (and thereby consistent with probabilistic logic), RELU is a better choice for activation functions. On the other hand, we were unable to show an inp-RELU-RELU network that converged to the copula representation $F_1$ .

Other networks such as inp-tanh-tanh learned the xor function for Boolean in-sample data but were unable to generalize to other real-valued xor representations. Of course, adding more records to the in-sample training set, increasing the number of inputs from two to three or more, and increasing the number of intermediate outputs (hidden layers), improves the approximation. But what is wrong with fixing the topology or activation function on a simpler implementation that allows less computational work, more accuracy, and faster convergence? Is there an "activation function bias" as well as data sample bias (for selecting both in-sample and out-sample data)? In this sense, we somewhat disagree with Hornick's observation [8] that

> …it's not the specific choice of the activation function, but rather the multilayer feedforward architecture itself which gives neural networks the potential of being universal learning machines.

Figure 7.2 shows an example of a cross-validation out-sample data set with a Boolean specification. Note that we identified three possible "acceptable" outputs (all shaded) corresponding to $s$=0, $s$=1, and $s$=∞.

| Binary in-Sample | x1 | x2 | targ | out-Sample | x1 | x2 | s=0? | s=1? | s=inf? |
|---|---|---|---|---|---|---|---|---|---|
| 0 | 0 | 0 | 0 | 0 | 0.5 | 1 | 0.5 | 0.5 | 0.5 |
| 1 | 0 | 1 | 1 | 1 | 0.5 | 0.5 | 0 | 0.5 | 1 |
| 2 | 1 | 0 | 1 | 2 | 0.75 | 0.25 | 0.5 | 0.625 | 1 |
| 3 | 1 | 1 | 0 | 3 | 0.75 | 0.5 | 0.25 | 0.5 | 0.75 |
| | | | | 4 | 0.75 | 0.75 | 0 | 0.375 | 0.5 |

**Figure 7.2. Boolean in-sample (4 records) and copula out-sample values (5 records) consistent with probabilistic logic (*s*=0, 1,∞).**

Again, we interpret numbers between zero and one as informal subjective probabilities. Can they be interpreted as actual probabilities? If the input and output probabilities are consistent with probabilistic logic, then it may be that the output corresponds to a copula representation of a Boolean formula.

Figure 7.3 shows four different in-sample training sets based on the above specifications. One specification each for Boolean, analog, copula; the fourth combines all three. Note that the analog and copula training sets are samples of an otherwise infinite set of values. For the combined set, only one input sample (0.5, 0.5, 0.5) corresponds to a copula specification ($s$=1). More in-sample data generally provides more opportunity for learning.

| Binary | | | |
|---|---|---|---|
| in-Sample | x1 | x2 | targ |
| 0 | 0 | 0 | 0 |
| 1 | 0 | 1 | 1 |
| 2 | 1 | 0 | 1 |
| 3 | 1 | 1 | 0 |

| Analog | | | | | | | |
|---|---|---|---|---|---|---|---|
| in-Sample | x1 | x2 | targ | in-Sample | x1 | x2 | targ |
| 0 | 0 | 0 | 0 | 5 | 1 | 0 | 1 |
| 1 | 0 | 0.5 | 0.5 | 6 | 1 | 0.5 | 0.5 |
| 2 | 0 | 0.75 | 0.75 | 7 | 1 | 0.75 | 0.25 |
| 3 | 0 | 1 | 1 | 8 | 1 | 1 | 0 |
| 4 | 0.5 | 0 | 0.5 | 9 | 0.5 | 1 | 0.5 |

| Copula | | | | | | | |
|---|---|---|---|---|---|---|---|
| in-Sample | x1 | x2 | targ | in-Sample | x1 | x2 | targ |
| 0 | 0.25 | 0.25 | 0.375 | 5 | 0.5 | 0.75 | 0.5 |
| 1 | 0.25 | 0.5 | 0.5 | 6 | 0.75 | 0.25 | 0.625 |
| 2 | 0.25 | 0.75 | 0.625 | 7 | 0.75 | 0.5 | 0.5 |
| 3 | 0.5 | 0.25 | 0.5 | 8 | 0.75 | 0.75 | 0.375 |
| 4 | 0.5 | 0.5 | 0.5 | | | | |

| All | | | | | | | |
|---|---|---|---|---|---|---|---|
| in-Sample | x1 | x2 | targ | in-Sample | x1 | x2 | targ |
| 0 | 0 | 0 | 0 | 5 | 0.5 | 1 | 0.5 |
| 1 | 0 | 0.5 | 0.5 | 6 | 1 | 0 | 1 |
| 2 | 0 | 1 | 1 | 7 | 1 | 0.5 | 0.5 |
| 3 | 0.5 | 0 | 0.5 | 8 | 1 | 1 | 0 |
| 4 | 0.5 | 0.5 | 0.5 | | | | |

**Figure 7.3. Boolean, Analog, Copula, and All (combined) in-sample training data consistent with probabilistic logic (*s*=1).**

## 7.2 Changing Topology: More Intermediate Outputs

In one experiment we doubled the intermediate outputs of the 2-2-1 network to see if the 2-4-1 network with inp-RELU-RELU can learn the *s*=1 copula. We used the "All" in-sample data set shown in Figure 7.2. The results, together with the $w^1_{11} \times w^1_{12}$ error surface projection (around the minimum) are shown in Figure 7.4: inp-RELU-RELU with 2-4-1 topology (with the "All" sample training set of Figure 7.3) can learn the *s*=1 representation. Figure 7.5 shows some results of the 2-4-1 network with inp-tanh-tanh with the "All" in-sample set for s=0 and s=1. The charts show that the results are approximating the *s*=0 and *s*=1 copulas. There were mixed results concerning the consistency with the probabilistic logic inequalities.

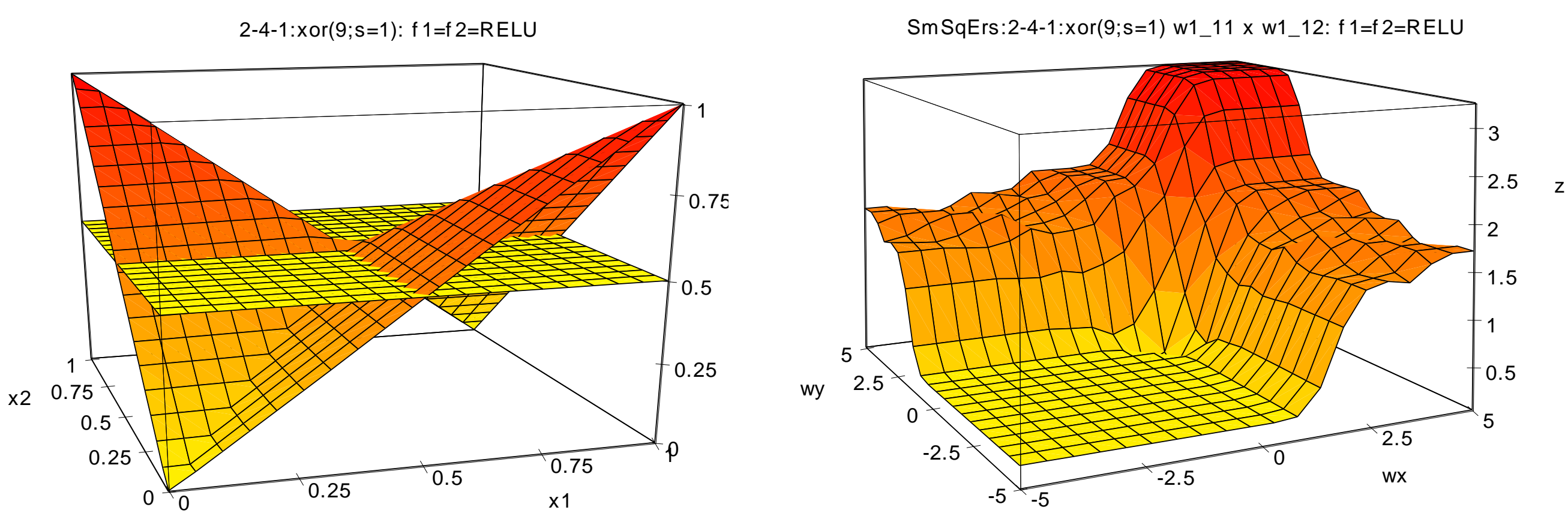

**Figure 7.4. Results of the 2-4-1 network with inp-RELU-RELU with the "All" 9-sample training set .**

**Figure 7.5. Results of the 2-4-1 network with inp-tanh-tanh .**

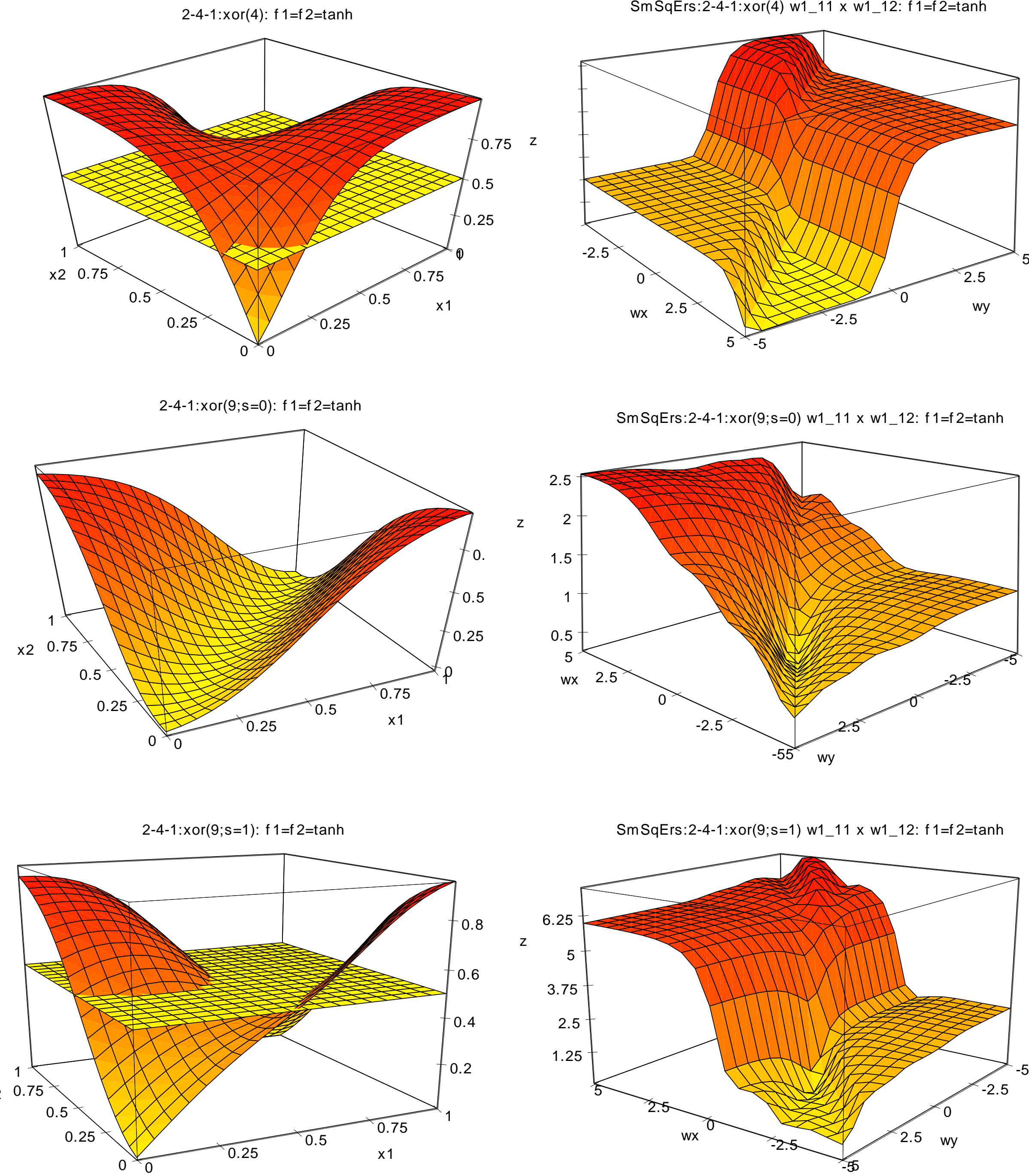

## 7.3 Changing Topology: More Hidden Layers

We observe that the more weights, the more complex the error surface in terms of folds and corners. In the last experiment we increased the number of layers. We kept the number of weights fixed. It is an exercise in accounting to verify that a 2-*n*-1 network has $4 \cdot n + 1$ weights; a 2-*p*-*q*-1 network has $p \cdot q + 3 \cdot p + 2 \cdot q + 1$ weights. Thus the 2-9-1 network (inp-tanh-tanh) and the 2-4-4-1 (inp-tanh-tanh-tanh) network have the same number of weights (37). We trained both networks on the 4-sample Boolean specification. Given similar in-sample error performance on the in-sample data, both networks converged to $u\left(\left|x_1 - x_2\right|\right)$ – the function observed in Section 5.1. Neither network was consistent with probabilistic logic. The 2-4-4-1 network converged in fewer iterations (having same goodness-of-fit error of less than 0.001) than the 2-9-1 network. Both networks are posted at [14].

| FINAL: 5000 iterates | | | | | | | | | | | | | | | | |
|---|---|---|---|---|---|---|---|---|---|---|---|---|---|---|---|---|
| | INP | w_1A | | (bias) | out_1A | | | | | | | | w_2A | | | outA |
| | 0.5 | 0.601 | 0.355 | -0.6 | -0.122 | 1.281 | 2.427 | 2.626 | 0.965 | 1.144 | 2.106 | 1.39 | 2.094 | 1.522 | 2.048 | 0.775 |
| | 0.5 | -0.28 | -0.98 | -0.48 | -0.803 | | | | | | | | | | | |
| | 1 | 2.612 | 2.53 | -0.64 | 0.959 | | | | | | | | | | | |
| | | 1.159 | 1.263 | -2.55 | -0.872 | | | | | | | | | | | |
| | | 0.321 | 0.297 | 1.509 | 0.949 | | | | | | | | | | | |
| | | -0.34 | -2.16 | 1.217 | -0.033 | | | | | | | | | | | |
| | | 0.404 | 0.725 | 1.026 | 0.92 | | | | | | | | | | | |
| | | -2.27 | 1.057 | -0.48 | -0.795 | | | | | | | | | | | |
| | | -1.17 | 0.133 | -0.58 | -0.8 | | | | | | | | | | | |
| | | | | | 1 | | | | | | | | | | | |

| FINAL: 1000 iterates | | | | | | | | | | | | | | | | | |
|---|---|---|---|---|---|---|---|---|---|---|---|---|---|---|---|---|---|
| | INP | w_1A | | | out_1A | | | w_2A | | | out_2A | | | w_3A | | | outA |
| | 0.5 | -0.8 | -0.53 | 0.69 | 0.029 | -1.072 | 0.256 | -0.87 | -1.04 | 0.236 | 0.897 | 0.627 | 0.247 | 1.671 | 0.982 | -0.75 | 0.959979 |
| | 0.5 | 0.897 | 1.025 | -0.99 | -0.025 | -0.607 | 0.433 | 0.5 | -0.46 | 1.212 | 0.755 | | | | | | |
| | 1 | -0.23 | -0.26 | -1 | -0.848 | 0.931 | -1.348 | -1.147 | 0.283 | -0.044 | 0.69 | | | | | | |
| | | -0.82 | -1.07 | 0.405 | -0.496 | -0.352 | 1.656 | 0.143 | -0.67 | 0.952 | 0.805 | | | | | | |
| | | | | | 1 | | | | | | 1 | | | | | | |

## 7.4 Lessons for larger networks?

We do not fully understand why the same 2-2-1 network trained with the same xor Boolean specification learns (when it converges) either $\mathrm{F}_0$ or $\mathrm{F}_\infty$. We did not see any other $\mathrm{F}_s$ besides the cases for $s = 0$ or $s = \infty$, even though by heuristic grounds we expected to see $\mathrm{F}_1$ in some of the training examples. Even though the error surface projections show the dynamics and problems associated with gradient search, the only conclusion we could make was that if the search starts from an unlucky point then the search will get stuck on a plateau.

The only other possibility could be related to the random sampling used in data selection. Selecting records of Boolean inputs corresponding to truth assignments could bias the underlying probabilistic logic statement probabilities. Would that be enough to influence the bifurcating xor convergence mechanism?

In any case: we can always evaluate a probability (consistent with probabilistic logic) of any Boolean expression formed with "and", "or", and "not" with $A(x, y)$ and $R(x, y)$. Since $A(x, y)$ and $R(x, y)$ are associative, this statement is still true for Boolean expressions of any number of inputs. Nonlinear networks with enough layers are universal approximators; they will converge to a function – even though (as we have seen with xor) the approximated function might not be the function that one wants.